\RequirePackage{silence}
\documentclass[sigconf, nonacm]{acmart}
\usepackage{enumitem}
\setlist{noitemsep,topsep=0pt,parsep=0pt,partopsep=0pt}

\providecommand{\deleted}[2][]{}

\providecommand{\definechangesauthor}[2][]{}
\makeatletter\@ifundefined{comment}{\newcommand{\comment}[2][]{}}{\renewcommand{\comment}[2][]{}}\makeatother

\makeatletter
\let\@sf@latexwarn\@latex@warning\let\@latex@warning\@gobble
\let\@sf@genericwarn\GenericWarning\let\GenericWarning\@gobbletwo
\makeatother
\usepackage{microtype}
\makeatletter
\let\@latex@warning\@sf@latexwarn
\let\GenericWarning\@sf@genericwarn
\makeatother
\usepackage{makecell}
\usepackage{booktabs}
\usepackage{algorithm}
\usepackage{algpseudocode}
\usepackage{subcaption}
\usepackage{flushend}
\usepackage[compact]{titlesec}
\titlespacing*{\section}{0pt}{6pt plus 3pt minus 1pt}{3pt plus 1pt}
\titlespacing*{\subsection}{0pt}{5pt plus 3pt minus 1pt}{2pt plus 1pt}
\theoremstyle{definition}
\newtheorem{definition}{Definition}

\AtBeginDocument{%
  \setlength{\abovedisplayskip}{3pt plus 1pt minus 1pt}%
  \setlength{\belowdisplayskip}{3pt plus 1pt minus 1pt}%
  \setlength{\abovedisplayshortskip}{0pt}%
  \setlength{\belowdisplayshortskip}{2pt plus 1pt minus 1pt}%
}

\begin{document}
\hypersetup{
  pdftitle={LLM-Guided ANN Index Optimization for Human-Object Interaction Retrieval},
  pdfauthor={Shahrzad Esmat, Chaunte W. Lacewell, Sameh Gobriel, Nilesh Jain, Ali Jannesari}
}
\title{LLM-Guided ANN Index Optimization for Human-Object Interaction Retrieval}

\author{Shahrzad Esmat}
\affiliation{%
  \institution{Iowa State University}
}
\email{sesmat@iastate.edu}

\author{Chaunté W. Lacewell}
\affiliation{%
  \institution{Intel Corporation}
}
\email{chaunte.w.lacewell@intel.com}

\author{Sameh Gobriel}
\affiliation{%
  \institution{Intel Corporation}
}
\email{sameh.gobriel@intel.com}

\author{Nilesh Jain}
\affiliation{%
  \institution{Intel Corporation}
}
\email{nilesh.jain@intel.com}

\author{Ali Jannesari}
\affiliation{%
  \institution{Iowa State University}
}
\email{jannesar@iastate.edu}

\begin{abstract}
Retrieval systems underpin modern AI applications --- spanning visual search,
recommendation engines, and multi-modal question answering.
Modern multi-stage retrieval systems require the joint optimization of highly coupled
parameters, yet traditional hyperparameter optimization (HPO) methods ---
including Tree-structured Parzen Estimators (TPE) and Gaussian Process Bayesian Optimization ---
rely on an independence assumption that fundamentally prevents them from
navigating these coupled configuration spaces.
We address this limitation with a phase-aware large language model (LLM) agent
that conditions each proposal on its full optimization history, navigating the
coupled parameter space across phase-partitioned exploration, exploitation, and
fine-tuning stages.
Evaluated on the HICO-DET human-object interaction retrieval benchmark using
Intel VDMS (Visual Data Management System), our agent outperforms Optuna~TPE
by $+33.3\%$ and VDTuner by $+34.2\%$ under SIEVE (Safeguarded Index Evaluation of Vector-search Efficiency, a quality-constrained
throughput metric), delivering a $15.3\times$ throughput gain over UniIR.
Validation across three benchmarks confirms that the agent's advantage grows with
the degree of parameter coupling: $+33.3\%$ on HICO-DET (high coupling),
methods converge within $1\%$ on GLDv2 (moderate coupling) and within $3.6\%$
on SIFT1M (near-independent control).
Cross-system validation on Milvus confirms the optimizer ranks first on all three
datasets without modification, demonstrating transferability across vector database
management system (VDBMS) platforms.
These results position our agent as the method of choice for retrieval systems
where parameter coupling is the dominant source of optimization complexity.
\end{abstract}

\maketitle

\pagestyle{plain}

\section{Introduction}
\label{sec:intro}

The throughput and accuracy of a vector database management system (VDBMS) depend
critically on approximate nearest neighbor (ANN) index configuration: on a
production-scale deployment, the gap between default and optimized settings can
exceed an \emph{order of magnitude} in queries per second on identical hardware.
These systems serve millions of queries daily across recommendation, visual
search, and cross-modal content discovery.
We study this problem in the context of Human-Object Interaction (HOI) retrieval
--- returning images of specific human-object interactions in response to natural
language queries --- an application with direct uses in video surveillance,
assistive robotics, and interactive multi-modal search.
Closing this performance gap demands an optimizer that treats parameters
jointly, not one that treats them as independent.

Modern vector databases such as VDMS (Visual Data Management System)~\cite{remis2021vdms},
Milvus~\cite{wang2021milvus}, and GaussDB-Vector~\cite{sun2025gaussdb} combine
two stages: a fast ANN search followed by a reranking step.
The parameters of each stage are tightly coupled: reducing retrieval search depth
(\textit{efSearch}) forces the reranker to compensate with a higher fusion
weight~($\alpha$) --- a cross-stage dependency that parameter-independent
optimizers cannot see.
On HICO-DET, the gap between an unoptimized baseline and our agent-optimized
configuration reaches $\mathbf{36.6\times}$ in throughput on a single A40 GPU
node --- the difference between 8 and 300 queries per second.
No existing method is designed to close this throughput gap.\looseness=-1

On the \emph{LLM-guided configuration} side, GPTuner~\cite{lao2024gpttuner}
and $\lambda$-Tune~\cite{giannakouris2025lambdatune} apply large language model (LLM) guidance to
database knob tuning, but their knobs are mutually independent --- no
cross-stage coupling exists.
On the \emph{ANN index tuning} side, VDTuner~\cite{yang2024vdtuner},
Optuna~TPE~\cite{akiba2019optuna}, grid and random
search~\cite{bergstra2012random} all treat parameters independently, missing
the coupling entirely.
VDTuner has a further weakness: it performs less reliably when performance
changes discontinuously at a quality threshold rather than gradually.\looseness=-1

We propose a \emph{phase-aware LLM agent} that jointly optimizes ANN index
and reranking parameters on Intel's VDMS~\cite{remis2021vdms}, and validate
that it transfers without modification to Milvus~\cite{wang2021milvus}.
At each step, the agent reviews all previous configurations and their results,
identifies how parameters across stages interact, and decides whether to explore
new regions or refine the best configuration found so far.
No existing benchmark combines fine-grained compositional text queries with a
large image corpus --- standard ANN benchmarks use synthetic vectors or
image-to-image similarity, while HOI datasets have only been used for
detection~\cite{kim2021hotr,zhang2021cdn,ning2023hoiclip,li2024dphoi}.
Alongside the agent, we introduce the first such benchmark built from
HICO-DET~\cite{chao2018hico}: its 600 verb--object categories
(e.g., \textit{ride horse}, \textit{hold umbrella}) map directly to
natural language queries, and its 90{,}641 annotated positive pairs provide
retrieval ground truth without additional labeling.\looseness=-1

Under SIEVE, our joint quality-throughput optimization objective defined in
Section~\ref{sec:metrics}, the agent achieves $\mathbf{+33.3\%}$ over
Optuna~TPE and $\mathbf{+34.2\%}$ over VDTuner~\cite{yang2024vdtuner} on HICO-DET.
On simpler datasets --- GLDv2~\cite{weyand2020gld} (image-to-image) and
SIFT1M~\cite{jegou2011pq} (pure ANN search) --- this advantage diminishes
proportionally to the degree of parameter coupling, confirming that LLM-guided
optimization is most valuable when parameters interact across stages.
Cross-system validation on Milvus~\cite{wang2021milvus} confirms the optimizer
transfers to a second VDBMS without modification, with LLM ranking first on all
three datasets.\looseness=-1

\noindent\textbf{Contributions.} This paper makes the following contributions:
\begin{itemize}[leftmargin=1em, itemsep=1pt, topsep=1pt]
  \item \textbf{Phase-aware LLM agent for coupled ANN index optimization.}
    We propose an LLM-guided optimizer that reads its own trial history to discover
    how index construction and reranking parameters must be configured together ---
    a dependency that standard optimizers miss by treating parameters independently.
    On HICO-DET the agent achieves $\mathbf{+33.3\%}$ over Optuna~TPE,
    $\mathbf{+34.2\%}$ over VDTuner~\cite{yang2024vdtuner}, and
    $\mathbf{+50.7\%}$ over GP-BO (Gaussian Process Bayesian Optimization)~\cite{snoek2012bayesian},
    with gains diminishing on simpler datasets where this inter-parameter dependency is absent.
    Cross-system transfer to Milvus~\cite{wang2021milvus} without modification confirms
    the improvement is structural, not incidental.

  \item \textbf{First text-query HOI retrieval benchmark.}
    Existing ANN benchmarks use synthetic vectors or image-to-image similarity, neither
    of which tests optimization under real semantic queries --- making it impossible to
    know whether smarter optimization actually matters in practice.
    We fill this gap by repurposing HICO-DET as a text-to-image retrieval challenge on
    a live vector database: 47{,}776 images queried by 600 natural-language verb--object
    phrases. The fine-grained compositional queries create the configuration-sensitive
    retrieval landscape that motivates our optimizer.

  \item \textbf{Single-index two-stage retrieval pipeline.}
    We present a two-stage pipeline that pairs Contrastive Language-Image
    Pre-training (CLIP)-based ANN search with DINOv2 reranking, gaining
    $\mathbf{+3.65}$\,pp mean Average Precision (mAP) at under 1\% extra latency.
    The design requires only a single ANN index, keeping the parameter space
    tractable for automated optimization while combining complementary visual
    and semantic cues.

  \item \textbf{Complexity-matched evaluation of when LLM reasoning pays off.}
    We evaluate across three datasets of decreasing complexity ---
    compositional HOI retrieval (HICO-DET), cross-domain landmark retrieval
    (GLDv2), and synthetic vector search (SIFT1M) --- and show that LLM-guided
    gains shrink as the search space becomes simpler, giving practitioners a
    clear signal for when the added complexity is worth it.

  \item \textbf{SIEVE: a quality-constrained throughput objective for
    deployment-aware ANN evaluation.}
    We propose SIEVE, a metric that enforces a minimum retrieval quality floor
    before rewarding throughput, matching the real deployment constraint that
    speed without accuracy is worthless.
    We show empirically that configurations ranked best by conventional smooth
    metrics routinely fall below this quality floor --- meaning standard metrics
    actively mislead optimization toward configurations that would fail in
    production.
\end{itemize}

\section{Related Work}
\label{sec:related}

\subsection{Vision-Language Retrieval and HOI Detection}
HOI detection has produced a series of strong transformer-based systems since
HICO-DET~\cite{chao2018hico} established the 600-category verb--object benchmark.
Human-Object interaction TRansformer (HOTR)~\cite{kim2021hotr} introduced end-to-end interaction detection via encoder-decoder
architectures, and Cascade Disentangling Network (CDN)~\cite{zhang2021cdn} disentangles human-object pair detection from interaction
classification in a cascade, extracting complementary strengths of one-stage and two-stage paradigms.
HOI-CLIP (HOICLIP)~\cite{ning2023hoiclip} showed that vision-language pre-training transfers
interaction knowledge efficiently, while Disentangled Pre-training for HOI (DPHoI)~\cite{li2024dphoi} exploited disentangled
pre-training over interaction primitives.
All operate in the detection paradigm --- predicting (person, object, interaction)
triples from an image --- and the sole prior non-detection use of
HICO-DET~\cite{kilickaya2021structured} employs spatial canvas composition with IoU
matching rather than free-form text queries.
Our novelty is specific to the intersection: general text-to-image retrieval benchmarks
(COCO Captions~\cite{chen2015coco}, Flickr30k~\cite{young2014flickr}) use open-domain descriptive captions; scene-graph retrieval
uses structured graph predicates; compositional image retrieval pairs images with
text modifications --- none provides free-form verb--object text queries against a
large HOI-annotated corpus with per-category retrieval ground truth, which is precisely
the combination that stresses ANN index configuration and motivates our optimizer.\looseness=-1

CLIP~\cite{radford2021clip} established joint vision-language embedding via contrastive
pre-training on 400 million image-text pairs, enabling zero-shot text-to-image retrieval.
DINOv2~\cite{oquab2024dinov2} produces self-supervised visual features that complement
CLIP's language-aligned representations, capturing fine-grained visual structure suited
to reranking.
UniIR~\cite{wei2024uniir} unified multimodal retrieval tasks under a single model
trained on M-BEIR, establishing a strong cross-modal baseline.
LamRA~\cite{liu2025lamra} proposed a unified framework in which a large multi-modal
model serves as both retriever and reranker across heterogeneous query types.
We fill the orthogonal HOI \emph{retrieval} direction and adopt CLIP and UniIR as
primary baselines.\looseness=-1

\subsection{Vector Database Systems and ANN Indexing}
The efficiency of HOI retrieval depends not only on the query models above but also on how the underlying approximate nearest-neighbor indices are configured.
ANN indexing algorithms span graph-based indices (e.g., HNSW~\cite{malkov2020hnsw} via
hierarchical navigable small-world graphs, DiskANN~\cite{subramanya2019diskann} for
billion-point SSD-resident datasets), inverted file structures with product
quantization~\cite{jegou2011pq}, and GPU-accelerated flat search~\cite{johnson2021billion}.
ANN-Benchmarks~\cite{aumuller2020ann} standardizes recall-versus-throughput evaluation
across these algorithms; the NeurIPS'21 competition~\cite{simhadri2022neurips} established
billion-scale baselines.\looseness=-1

Production VDBMSs integrate ANN indexing with metadata management and query processing.
Milvus~\cite{wang2021milvus} supports multiple index backends with a cloud-native query
layer.
GaussDB-Vector~\cite{sun2025gaussdb} separates adjacency-list from vector storage and
applies novel buffering strategies to sustain low-latency high-recall search at billion-vector scale.
VSAG~\cite{zhong2025vsag} exploits a subgraph inclusion property --- stricter-parameter
graphs are subgraphs of relaxed-parameter ones --- enabling parameter tuning without index rebuilds.
VDMS~\cite{remis2021vdms} co-locates graph-property metadata with vector data and exposes
HNSW index parameters through a Python API, making it the natural platform for an
optimizer that must rebuild and re-query the index at each evaluation step.
Across all these systems, index configuration is left to manual tuning or blind search;
automated optimization for multi-stage coupled parameter spaces remains an open problem.\looseness=-1

\subsection{Hyperparameter Optimization}
Standard hyperparameter optimization (HPO) methods include random search~\cite{bergstra2012random}, which provides a
strong baseline under bounded evaluation budgets; Bayesian optimization with Gaussian
Process surrogates~\cite{snoek2012bayesian}, which applies acquisition functions to
balance exploration and exploitation; and bandit-based scheduling such as
Hyperband~\cite{li2018hyperband}, which allocates evaluation budget adaptively.
Auto-sklearn~\cite{feurer2015autosklearn} combines Bayesian optimization with
meta-learning over prior tasks to warm-start configuration search.
Optuna~\cite{akiba2019optuna} implements Tree-structured Parzen Estimators for
mixed-type spaces and serves as our strongest non-LLM baseline.\looseness=-1

For database management system (DBMS) tuning, OtterTune~\cite{vanaken2017ottertune} pioneered ML-guided knob
selection via Gaussian Process regression over workload representations.
VDTuner~\cite{yang2024vdtuner} extends this approach to multi-objective Bayesian
optimization across seven index types in Milvus, improving over manual baselines but
operating on single-stage vector search without cross-stage coupling.
All of these methods --- standard HPO~\cite{bergstra2012random,snoek2012bayesian,li2018hyperband,akiba2019optuna}, AutoML~\cite{feurer2015autosklearn}, and system-specific Bayesian optimization~\cite{vanaken2017ottertune,yang2024vdtuner}
--- are designed for single-stage configuration spaces; none addresses the compensatory
interactions that arise when index and reranking parameters are coupled across retrieval stages.\looseness=-1

\subsection{LLM-Guided System Configuration}
Recent work has explored whether large language models can serve as optimization agents, bringing natural language reasoning to configuration search.
OPRO~\cite{yang2024opro} established that LLMs can act as black-box optimizers by
conditioning successive proposals on a history of candidate solutions and their scores.
Follow-up work extended this paradigm to ML hyperparameter
optimization~\cite{zhang2023llmhyperopt,liu2024agenthpo}.\looseness=-1

GPTuner~\cite{lao2024gpttuner} applies LLM-guided Bayesian optimization to PostgreSQL
knob tuning, extracting structured configuration knowledge from database documentation
to guide search.
$\lambda$-Tune~\cite{giannakouris2025lambdatune} generates complete database
configuration scripts via LLM, framing prompt construction as a cost-based optimization,
and is evaluated across PostgreSQL and MySQL.
Both outperform ML-based and RL-based baselines on DBMS tuning, yet both operate on
flat, independently interpretable knob spaces where no cross-stage coupling exists by
construction.\looseness=-1

Our work addresses what neither thread has reached: a coupled multi-stage parameter
space where ANN retrieval depth and fusion reranking weight must be optimized jointly
under a hard feasibility constraint that smooth surrogate models cannot represent.
Three structural properties make directly applying OPRO~\cite{yang2024opro} insufficient.
First, the valid parameter set is \emph{conditional} on engine type --- the search space
itself changes mid-optimization, violating OPRO's fixed-candidate assumption.
Second, the SIEVE feasibility cliff is discontinuous: plain history-conditioned proposals
average over the cliff rather than committing to the feasible high-QPS side.
Third, cross-stage $(k,\alpha)$ coupling means the joint optimum is unreachable by
treating parameters as independent marginals, as history-only proposals do.
Our additions --- phase structure, diagnostic guidance $g(\mathcal{H})$, and
untried-value hints --- directly target each gap (Section~\ref{sec:ablation}
brackets OPRO: removing history gives $\mathbf{0/3}$ seeds; retaining history
without phases/hints upper-bounds OPRO at $299.5$, $\mathbf{3/3}$ seeds).\looseness=-1

\section{The HICO-DET Retrieval Benchmark}
\label{sec:benchmark}

\subsection{Source Dataset}
HICO-DET~\cite{chao2018hico} contains 47{,}776 images (38{,}118 train / 9{,}658 test)
annotated across 600 HOI categories with per-instance bounding boxes and interaction
labels, where each category is a \emph{verb--object} pair drawn from 117 action verbs
and 80 COCO object classes.
Three structural properties make HICO-DET particularly challenging for ANN index
configuration optimization.
First, \emph{compositional fine-grained discrimination}: the 600 categories span
80 COCO object classes and 117 verb types; categories sharing the same object
(e.g., \textit{ride horse}, \textit{sit\_on horse}, \textit{straddle horse})
differ only in the action verb, so single-encoder retrieval conflates them and
the joint selection of ANN engine and Stage-2 fusion weight becomes critical.
Second, \emph{long-tail relevance}: category sizes are highly skewed, with the
largest exceeding 1{,}000 positive images and the smallest having as few as 5;
an optimizer that maximizes mean mAP without attending to per-category variance
will exploit high-frequency categories while failing on rare but equally weighted
queries.
Third, a \emph{dual-encoder modality gap} requires joint tuning: CLIP retrieves
by language-aligned semantic similarity while DINOv2 captures fine-grained visual
structure complementary to CLIP's representations~\cite{oquab2024dinov2}; CLIP
alone (Stage~1 without DINOv2 reranking) achieves lower mAP than the two-stage
fusion, so the Stage-2 fusion weight $\alpha$ is a data-dependent parameter that
must be tuned jointly with the Stage-1 ANN configuration.\looseness=-1

\subsection{Benchmark Construction}
\noindent\textbf{Retrieval Corpus.}
We use all 47{,}776 images as a single retrieval corpus.
The original train/test split was designed for supervised detection training and has
no meaning for retrieval evaluation, where the task is to rank corpus images given a
query; merging both splits maximizes the per-category positive density needed for
reliable mAP computation across all 600 categories.
CLIP ViT-L/14~\cite{radford2021clip} embeddings are pre-computed offline using the
official OpenAI weights and stored in VDMS as a 768-dimensional L2 DescriptorSet
of 47{,}776 vectors.\looseness=-1

\noindent\textbf{Query Design: One Canonical Query per HOI Class.}
We construct exactly one canonical natural-language text query per HOI category,
yielding 600 queries.
A single canonical query per category eliminates inter-query variance from paraphrase
choice and ensures that each of the 600 HICO-DET categories contributes equally to
the mean Average Precision score~\cite{manning2008ir}.
Each query follows the template \textit{``a person \{verb\} a \{object\}''}, with the
special case \textit{``a person near a \{object\}''} for the \texttt{no\_interaction}
verb class.
Queries are encoded with the same CLIP ViT-L/14 text encoder, producing
768-dimensional unit-norm text embeddings that reside in the same embedding space as
the indexed image vectors.\looseness=-1

\noindent\textbf{Ground Truth: Multi-Label Positive Structure.}
An image is relevant to a query if it carries the corresponding HOI annotation in
the HICO-DET ground truth.
A single image may be relevant to multiple queries (e.g., an image annotated with both
\textit{ride motorcycle} and \textit{hold phone} contributes positives to two queries).
Summed over all 600 queries, there are 90{,}641 positive (image, query) pairs.
For each HOI category, the single image used as the DINOv2 reranking reference
is excluded from relevance judgments (treated as a junk image), ensuring the
reference cannot trivially self-retrieve; after this exclusion the minimum
per-category positive count is 5, so mAP is well-defined for all 600 queries.\looseness=-1

\subsection{Evaluation Protocol}
We evaluate retrieval quality using mean Average Precision (mAP),
Precision@10, Normalized Discounted Cumulative Gain (NDCG)@10, and Recall@10 over all 600 queries.
mAP is our primary metric: it rewards systems that rank relevant images
higher in the retrieved list and is the standard metric for ranked retrieval
with multiple relevant documents per query~\cite{manning2008ir}.
Throughput is measured as batch QPS: 600 queries issued as a single batch to a
live VDMS container, divided by total wall-clock time.\looseness=-1
The joint optimization objective is
\begin{equation}
  \mathrm{Score}(\theta) =
  \begin{cases}
    \mathrm{QPS}(\theta) & \text{if } \mathrm{mAP}(\theta) \geq \tau, \\
    0                    & \text{otherwise,}
  \end{cases}
  \quad \tau = 0.15 .
  \label{eq:score}
\end{equation}
This threshold-gated objective rewards throughput only within the feasible region
($\mathrm{mAP} \geq \tau$): a configuration that sacrifices retrieval quality below
$\tau$ scores zero regardless of how high its QPS is, while within the feasible
region higher QPS is strictly better.
$\tau{=}0.15$ is calibrated as the minimum acceptable quality floor: it lies between
UniIR's mAP ($0.095$, failing) and the unoptimized CLIP baseline ($0.208$, passing),
ensuring optimized configurations retain at least the quality of the simplest
non-trivial retrieval system; rankings are stable for $\tau\!\in\![0.15,0.20]$,
and below $0.15$, $k{=}50$ configurations with mAP\,$\approx\,0.13$ --- which found high
throughput but lack the compensating $\alpha{=}0.80$ --- become spuriously feasible.\looseness=-1

\subsection{Problem Statement}
\label{sec:probstmt}

\begin{definition}[Configuration Space]\label{def:space}
Let $\Theta$ denote the joint configuration space of the two-stage pipeline
(Section~\ref{sec:pipeline}).
$\Theta$ is the Cartesian product of seven parameter domains
(\emph{engine}; $M$, $\mathit{nlist}$; $\mathit{efSearch}$, $\mathit{nprobe}$;
$k$; $\alpha$; $n_{\mathrm{refs}}$; $\mathit{ref\_strategy}$)
enumerated in Table~\ref{tab:config-space}; $|\Theta| = 102{,}144$.\footnote{The LLM agent also proposes a \texttt{constraint\_strategy}
  metadata pre-filter ($\in$\{\texttt{none, object, verb, object\_and\_verb}\}) outside $\Theta$;
  all methods sample the full space, but the LLM converges to \texttt{object} in 66\% of trials
  and uses it in all best configurations.}
\end{definition}

\begin{definition}[Evaluation Oracle]\label{def:oracle}
For any configuration $\theta \in \Theta$, the evaluation oracle
$\mathcal{O}(\theta) = \bigl(\mathrm{mAP}(\theta),\,\mathrm{QPS}(\theta)\bigr)$
is obtained by (i)~rebuilding the VDMS ANN index under $\theta$, and
(ii)~issuing all 600 HICO-DET text queries to the live container and recording
mAP (Section~\ref{sec:benchmark}) and batch QPS.
The oracle is \emph{black-box}: no closed-form gradient with respect to
$\theta$ is available, and each call requires a full index rebuild followed by
600 retrieval operations.
\end{definition}

\begin{definition}[Optimization Problem]\label{def:optproblem}
Given $\Theta$ (Definition~\ref{def:space}), $\mathcal{O}$ (Definition~\ref{def:oracle}),
and a finite evaluation budget $N \in \mathbb{N}$, find
\begin{equation}
  \theta^{*} \;=\; \arg\max_{\theta \in \Theta}\;
  \mathrm{Score}(\theta) ,
  \label{eq:optproblem}
\end{equation}
subject to at most $N$ calls to $\mathcal{O}$, where $\mathrm{Score}(\theta)$ is
defined by Eq.~\eqref{eq:score}
($\mathrm{QPS}(\theta)$ if $\mathrm{mAP}(\theta) \geq \tau$, else $0$).
We set $N{=}50$ throughout, covering $0.05\%$ of $|\Theta|$.
\end{definition}

\begin{table}[t]
\centering
\caption{Configuration space $\Theta$ (Definition~\ref{def:space}); $^\dagger$conditional on engine; total $|\Theta|{=}102{,}144$.}
\label{tab:config-space}
\resizebox{\columnwidth}{!}{%
\setlength{\tabcolsep}{4pt}
\begin{tabular}{@{}llp{3.6cm}r@{}}
\toprule
\textbf{Parameter} & \textbf{Notation} & \textbf{Value Set} & \textbf{Card.} \\
\midrule
\multicolumn{4}{@{}l}{\textit{Stage 1 --- ANN retrieval}} \\
\quad engine
  & ---
  & FaissFlat, FaissHNSWFlat, FaissIVFFlat
  & 3  \\
\quad conn.\ depth$^\dagger$\,(HNSW)
  & $M$
  & 8, 16, 32, 48, 64
  & 5  \\
\quad cluster count$^\dagger$\,(IVF)
  & $\mathit{nlist}$
  & 64, 128, 256, 512
  & 4  \\
\quad search beam$^\dagger$\,(HNSW)
  & $\mathit{efSearch}$
  & 32--500\,(7 vals)
  & 7  \\
\quad probed cells$^\dagger$\,(IVF)
  & $\mathit{nprobe}$
  & 4, 8, 16, 32, 64
  & 5  \\
\quad candidate pool
  & $k$
  & 50--1000\,(8 vals)
  & 8  \\
\midrule
\multicolumn{4}{@{}l}{\textit{Stage 2 --- DINOv2 reranking}} \\
\quad fusion weight
  & $\alpha$
  & 0.0, 0.05, \ldots, 0.90
  & 19 \\
\quad ref.\ count
  & $n_{\mathrm{refs}}$
  & 1, 3, 5, 10
  & 4  \\
\quad ref.\ strategy
  & ---
  & first, centroid, diverse
  & 3  \\
\midrule
\multicolumn{3}{@{}l}{\textbf{Total} $|\Theta|$}
  & \textbf{102,144} \\
\bottomrule
\end{tabular}%
}
\end{table}

\section{Two-Stage Retrieval Pipeline}
\label{sec:pipeline}

\begin{figure*}[t]
  \centering
  \includegraphics[width=0.76\textwidth]{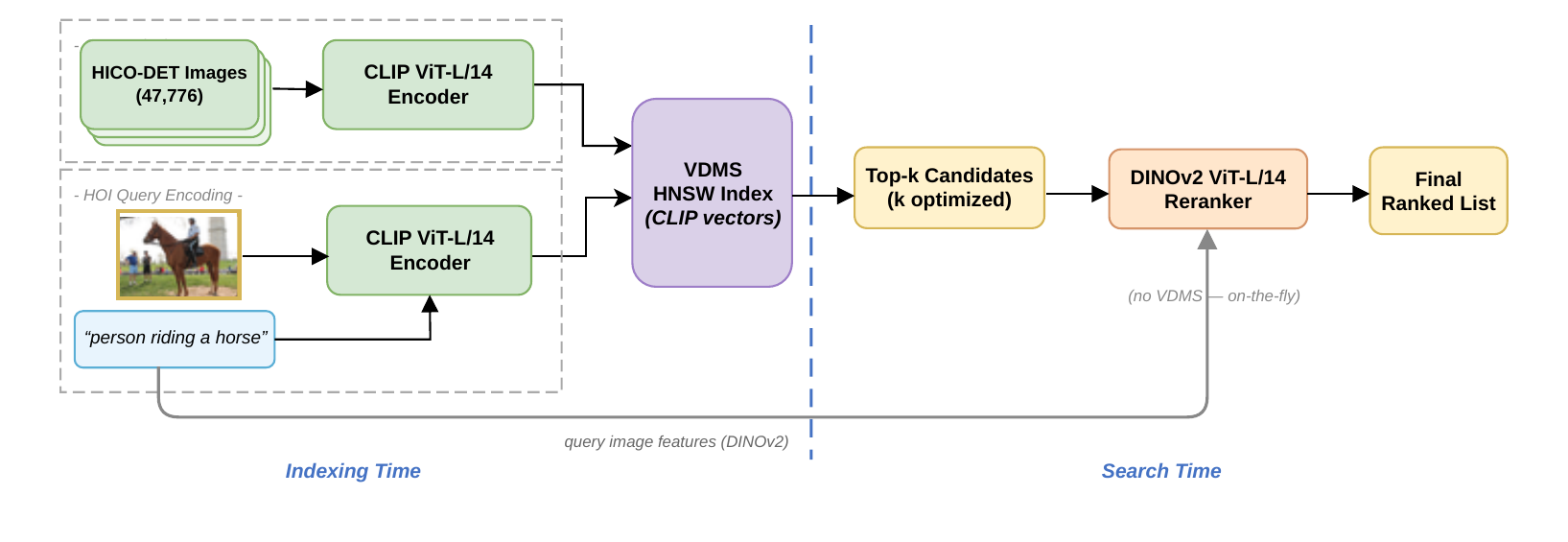}
  \Description{Two-stage retrieval pipeline diagram.}
  \caption{Two-stage retrieval pipeline (CLIP ANN retrieval + DINOv2 reranking).}
  \label{fig:pipeline}
\end{figure*}

\subsection{Oracle Instantiation and Notation}

Oracle $\mathcal{O}$ (Definition~\ref{def:oracle}) maps any $\theta \in \Theta$ to a
$(\mathrm{mAP},\,\mathrm{QPS})$ pair; the optimizer maximizes
$\mathrm{Score}(\theta)$ (Eq.~\eqref{eq:score}): $\mathrm{QPS}(\theta)$ if
$\mathrm{mAP}(\theta) \geq \tau$, else $0$.
Let $\mathcal{C} = \{(\mathbf{c}_i,\,\mathbf{d}_i)\}_{i=1}^{N}$ denote the corpus
of $N{=}47{,}776$ images, where $\mathbf{c}_i \in \mathbb{R}^{768}$ is the
unit-normalized CLIP ViT-L/14 embedding and $\mathbf{d}_i \in \mathbb{R}^{1024}$
is the DINOv2 ViT-L/14-reg4 embedding of image $i$,
and $\mathcal{I}_h \subseteq [N]$ the annotated relevant set for HOI class $h$.
Given a text query $q_h$ (e.g., \textit{``a person riding a horse''}), the pipeline
proceeds in two stages as illustrated in Figure~\ref{fig:pipeline}.

\textbf{Stage~1} encodes $q_h$ into a unit-norm vector $\mathbf{q}_c \in \mathbb{R}^{768}$
via the frozen CLIP ViT-L/14 text encoder and issues an ANN search over the VDMS
CLIP DescriptorSet to retrieve a candidate pool $\mathcal{T}_k(q_h)$ of cardinality $k$.
\textbf{Stage~2} reranks $\mathcal{T}_k(q_h)$ using DINOv2 visual distances computed
against a pre-built per-class reference $\mathbf{r}_h$, fusing both distance signals
without any additional index.
The seven parameters of $\theta$ --- engine, $M$ or $\mathit{nlist}$,
$\mathit{efSearch}$ or $\mathit{nprobe}$, $k$, $\alpha$, $n_{\mathrm{refs}}$,
$\mathit{ref\_strategy}$ --- govern the accuracy--throughput trade-off across both
stages and constitute the search space $\Theta$ (Table~\ref{tab:config-space}).
The oracle cost per call is dominated by index construction (${\approx}425$--$455$~s)
and 600 batch queries (${\approx}35$--$99$~s), yielding ${\approx}8$--$9$~minutes
of wall-clock time per oracle evaluation.\looseness=-1

\subsection{Stage 1: ANN Candidate Generation}
\label{sec:stage1}

All $N$ CLIP ViT-L/14 image embeddings are pre-computed offline and stored in VDMS as
a 768-dimensional L2 DescriptorSet.
At query time, $q_h$ is encoded by the frozen text encoder and submitted to VDMS
as a \texttt{FindDescriptor} request.
VDMS returns the top-$k$ candidate identifiers together with their squared L2 distances
$\delta_j^c = \|\mathbf{c}_j - \mathbf{q}_c\|_2^2$; the candidate pool is formally
\begin{equation}
  \mathcal{T}_k(q_h) \;=\; \bigl\{j \in [N] :\,\|\mathbf{c}_j - \mathbf{q}_c\|_2^2
    \leq \delta^{(k)}\bigr\} ,
  \label{eq:stage1}
\end{equation}
where $\delta^{(k)}$ denotes the $k$-th smallest squared L2 distance from $\mathbf{q}_c$ to
any corpus embedding.
For FaissFlat, Eq.~\eqref{eq:stage1} is satisfied exactly; for FaissHNSWFlat and
FaissIVFFlat, VDMS returns a high-recall approximation whose tightness is governed
by the engine-specific search parameters.

Three ANN engines expose qualitatively distinct accuracy--throughput operating points.

\noindent\textbf{FaissFlat:} exact exhaustive L2 search; recall upper bound at lowest QPS.

\noindent\textbf{FaissHNSWFlat}~\cite{malkov2020hnsw}: hierarchical navigable small-world graph; recall/latency controlled by graph connectivity $M$ and beam width $\mathit{efSearch}$ (values in Table~\ref{tab:config-space}).
The build-quality parameter $\mathit{efConstruction}$ is fixed at $200$ throughout HICO-DET and GLDv2 experiments; preliminary experiments confirmed that values above $200$ yield negligible recall improvement at substantially higher build cost, so it is excluded from $\Theta$.

\noindent\textbf{FaissIVFFlat}~\cite{johnson2021billion}: inverted-file index; search probes $\mathit{nprobe}$ of $\mathit{nlist}$ Voronoi cells (values in Table~\ref{tab:config-space}).

The candidate pool size $k$ (8 values: 50--1000, Table~\ref{tab:config-space}) determines
the recall ceiling available to Stage~2 and is selected jointly with engine parameters.
Batch QPS is defined as
\begin{equation}
  \mathrm{QPS}(\theta) \;=\; \frac{|Q|}{T_{\mathrm{total}}(\theta)} ,
  \label{eq:qps}
\end{equation}
where $|Q|{=}600$ is the number of HOI text queries and $T_{\mathrm{total}}(\theta)$
is the elapsed wall-clock time for all queries issued in a single batch to the live
VDMS container under configuration $\theta$.
This end-to-end latency encompasses the VDMS round-trip for Stage~1, the in-memory
reranking computation in Stage~2, and all inter-process communication overhead.\looseness=-1

\subsection{Stage 2: Feature Fusion and Reranking}
\label{sec:stage2}

Stage~2 reranks $\mathcal{T}_k(q_h)$ using DINOv2 ViT-L/14-reg4~\cite{oquab2024dinov2}
visual features --- capturing fine-grained appearance structure complementary to
CLIP's language-aligned semantics --- without constructing any additional index.
Each HOI class $h$ is represented by a visual reference vector
$\mathbf{r}_h \in \mathbb{R}^{1024}$, built offline as the mean DINOv2 embedding
of a selected subset $\mathcal{S}_h \subseteq \mathcal{I}_h$:
\begin{equation}
  \mathbf{r}_h \;=\; \frac{1}{|\mathcal{S}_h|} \sum_{i \in \mathcal{S}_h} \mathbf{d}_i ,
  \label{eq:refvec}
\end{equation}
where $n_r = |\mathcal{S}_h| \in \{1, 3, 5, 10\}$.
For $n_r{=}1$ (the default), $\mathbf{r}_h$ is the DINOv2 embedding of a single image
from the HOI category.
For $n_r{>}1$, three strategies populate $\mathcal{S}_h$:
\textbf{first} takes the initial $n_r$ images in dataset order;
\textbf{centroid} selects the $n_r$ images nearest to the DINOv2 class mean
$\boldsymbol{\mu}_h^d = \frac{1}{|\mathcal{I}_h|}\sum_{i \in \mathcal{I}_h}\mathbf{d}_i$,
anchoring $\mathbf{r}_h$ at the visual core of the class;
\textbf{diverse} applies greedy farthest-point sampling from the most central image.
Reference construction requires ${<}1$\,ms per class and does not contribute to
$T_{\mathrm{total}}$ in Eq.~\eqref{eq:qps}.\looseness=-1

CLIP squared L2 distances $\delta_j^c = \|\mathbf{c}_j - \mathbf{q}_c\|_2^2$ and DINOv2 L2 distances
$\delta_j^d = \|\mathbf{d}_j - \mathbf{r}_h\|_2$ operate at different absolute scales;
both are min-max normalized within $\mathcal{T}_k(q_h)$ onto $[0,\,1]$:
\begin{equation}
  \hat{\delta}_j^m \;=\;
    \frac{\delta_j^m - \min_{i \in \mathcal{T}_k}\delta_i^m}
         {\max_{i \in \mathcal{T}_k}\delta_i^m
          - \min_{i \in \mathcal{T}_k}\delta_i^m + \varepsilon} ,
  \quad m \in \{c,\,d\},
  \label{eq:norm}
\end{equation}
where $\varepsilon = 10^{-10}$, $\hat{c}_j \equiv \hat{\delta}_j^c$, and
$\hat{d}_j \equiv \hat{\delta}_j^d$.
Pool-local normalization preserves the fine-grained relative ordering of CLIP distances
within $\mathcal{T}_k(q_h)$: intra-pool CLIP squared L2 distances span a narrow range
(${\approx}1.55$--$1.75$ for unit-normalized 768-d embeddings), a signal that a
global corpus statistic would collapse.\looseness=-1

The fused relevance score for candidate $j \in \mathcal{T}_k(q_h)$ is the
convex combination~\cite{bruch2024tois}
\begin{equation}
  s_j \;=\; (1-\alpha)\,\hat{c}_j \;+\; \alpha\,\hat{d}_j ,
  \label{eq:fusion}
\end{equation}
where $\alpha \in \{0.0, 0.05, \ldots, 0.90\}$ (19 values); lower $s_j$ indicates
higher predicted relevance.
Setting $\alpha{=}0$ degenerates to CLIP-only retrieval; $\alpha{=}1$ to
DINOv2-only reranking.
The convex combination is preferred over reciprocal rank fusion because CLIP text
queries on a unit-normalized 768-d space produce meaningful intra-pool distance values
whose fine-grained ordering is destroyed when converted to ranks.
The optimal $\alpha$ is configuration-dependent and is learned jointly with Stage~1
parameters by the LLM optimizer (Section~\ref{sec:optimizer}).\looseness=-1

\subsection{Text Query Superiority: Centroid-Displacement Theory}
\label{sec:centroid}

Let $\boldsymbol{\mu}_h^c = \frac{1}{|\mathcal{I}_h|}\sum_{i\in\mathcal{I}_h}\mathbf{c}_i$
denote the visual centroid of HOI class $h$ in the CLIP space, and let
$\sigma_h^2 = \frac{1}{|\mathcal{I}_h|}\sum_{i \in \mathcal{I}_h}
\|\mathbf{c}_i - \boldsymbol{\mu}_h^c\|_2^2$ denote the intra-class visual variance.
For any query vector $\mathbf{q}' \in \mathbb{R}^{768}$, the bias--variance
decomposition of the expected squared L2 distance to a class member $\mathbf{c}_i$
drawn uniformly from $\mathcal{I}_h$ yields
\begin{equation}
  \mathbb{E}_{i}\bigl[\|\mathbf{q}' - \mathbf{c}_i\|_2^2\bigr]
  \;=\;
  \underbrace{\|\mathbf{q}' - \boldsymbol{\mu}_h^c\|_2^2}_{\text{centroid bias}^2}
  \;+\;
  \underbrace{\sigma_h^2}_{\text{intra-class variance}} .
  \label{eq:biasvar}
\end{equation}
The variance term $\sigma_h^2$ is an intrinsic property of the class; only the
bias term $\|\mathbf{q}' - \boldsymbol{\mu}_h^c\|_2^2$ depends on the query.

CLIP's contrastive training objective aligns the text embedding of a category
description with the visual embeddings of all matching images~\cite{radford2021clip}.
For a compositionally structured category --- e.g., \textit{``a person riding a horse''}
spanning diverse viewpoints, backgrounds, and subjects --- the text embedding
aggregates distributional statistics from many matched images and therefore
lies near the visual centroid: $\mathbf{q}_c \approx \boldsymbol{\mu}_h^c$.
The centroid bias is thus approximately zero:
\begin{equation}
  \mathbb{E}_{i}\bigl[\|\mathbf{q}_c - \mathbf{c}_i\|_2^2\bigr]
  \;\approx\; \sigma_h^2 .
  \label{eq:textquery}
\end{equation}
In contrast, any reference image $\mathbf{c}_j$ ($j \in \mathcal{I}_h$) lies at
centroid displacement $\varepsilon_j = \|\mathbf{c}_j - \boldsymbol{\mu}_h^c\|_2 > 0$,
incurring a strictly positive bias:
\begin{equation}
  \mathbb{E}_{i}\bigl[\|\mathbf{c}_j - \mathbf{c}_i\|_2^2\bigr]
  \;=\; \varepsilon_j^2 + \sigma_h^2
  \;>\; \sigma_h^2 .
  \label{eq:imagequery}
\end{equation}
The text query therefore achieves lower expected squared distance to every
class member than any single reference image, by a margin that shrinks as
$\mathbf{q}_c$ approaches $\boldsymbol{\mu}_h^c$.
This centroid-displacement advantage is monotonically amplified by intra-class visual
diversity: the higher the variance $\sigma_h^2$ across HOI images (e.g., \textit{ride
horse} spans many settings and viewpoints), the larger the typical $\varepsilon_j$
and hence the larger the margin over any individual reference image.
Under L2-based ANN retrieval, lower expected squared distance to class members
translates directly to higher expected recall at any fixed $k$, and therefore to
higher mAP.
This result formally justifies the design choice of text queries for Stage~1 and
additionally motivates the $n_r{>}1$ reference aggregation in Stage~2: averaging
$n_r$ randomly drawn class embeddings reduces the expected $L_2$ centroid bias of $\mathbf{r}_h$
by $O(1/\sqrt{n_r})$ relative to a single reference image.
Section~\ref{sec:experiments} confirms this empirically: text queries outperform
single-image reference queries by $+2.32$\,pp mAP on HICO-DET, consistent with
Eq.~\eqref{eq:textquery}.\looseness=-1

\section{LLM-Guided ANN Hyperparameter Optimization}
\label{sec:optimizer}

\begin{figure*}[t]
  \centering
  \includegraphics[width=0.92\textwidth, trim=0 102 10 0, clip]{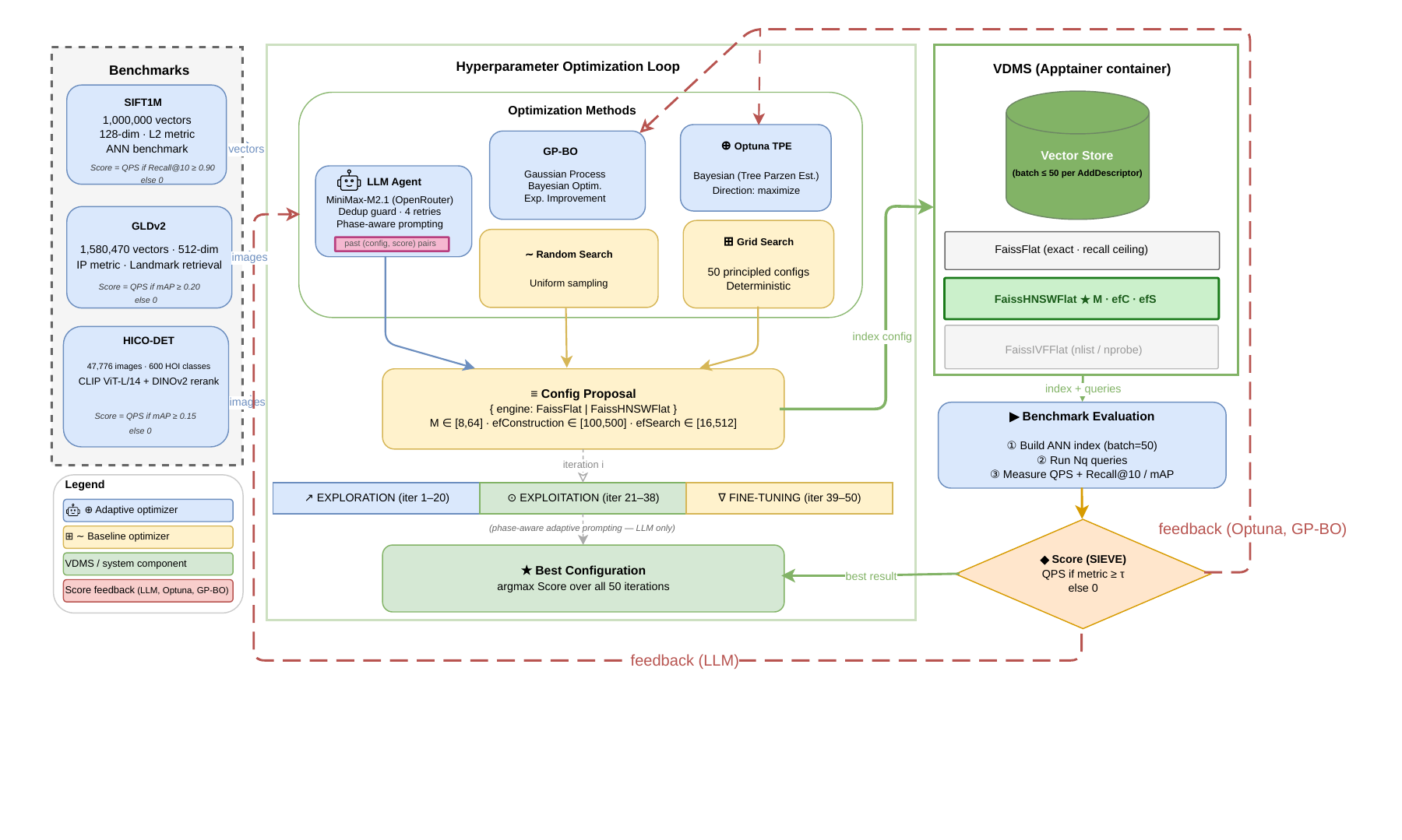}
  \Description{Optimization framework diagram showing five methods.}
  \caption{Optimization framework: five methods over $N{=}50$ iterations (adaptive methods receive score feedback).}
  \label{fig:workflow}
\end{figure*}

The pipeline of Section~\ref{sec:pipeline} exposes a seven-parameter joint configuration
$\theta \in \Theta$: engine, two engine-conditional index parameters, pool size $k$,
fusion weight $\alpha$, reference count $n_r$, and reference strategy.
Because engine choice renders some parameters inapplicable---$M$ and
$\mathit{efSearch}$ are undefined for IVF; $\mathit{nlist}$ and $\mathit{nprobe}$
are undefined for HNSW---the joint space $|\Theta|{=}102{,}144$ is
engine-conditional and mixed-type.\looseness=-1

Each evaluation requires a full VDMS rebuild plus a 600-query pass
(${\approx}8$--$9$ min per oracle call), ruling out exhaustive or
surrogate-heavy strategies that require thousands of evaluations.
We address this with a sequential LLM optimizer (Figure~\ref{fig:workflow}) that exploits structural knowledge
of $\Theta$ to allocate its $N$-iteration budget efficiently.\looseness=-1

\subsection{Search Space and Coupling Structure}

The configuration space decomposes as
\begin{equation}
  \Theta \;=\; \Theta_{\mathrm{HNSW}} \;\cup\; \Theta_{\mathrm{IVF}}
               \;\cup\; \Theta_{\mathrm{Flat}},
\end{equation}
where each component is defined by its engine-conditional index parameters together
with shared parameters $k$, $\alpha$, $n_r$, and $\mathrm{ref\_strategy}$
(full value sets in Table~\ref{tab:config-space}).
FaissHNSWFlat contributes $M$ and $\mathit{efSearch}$;
FaissIVFFlat contributes $\mathit{nlist}$ and $\mathit{nprobe}$.
The oracle $\mathcal{O}(\theta)$ loads $\theta$ into a fresh VDMS instance,
indexes the corpus, and evaluates all $|Q|{=}600$ HOI queries, returning the pair
$\bigl(\mathrm{mAP}(\theta),\mathrm{QPS}(\theta)\bigr)$
and $\mathrm{Score}(\theta)=\mathrm{QPS}(\theta)$ if $\mathrm{mAP}(\theta){\geq}\tau$, else $0$ (Eq.~\eqref{eq:score}).\looseness=-1

Two cross-parameter couplings invalidate the independence assumption shared by
GP-BO~\cite{snoek2012bayesian} and Optuna~TPE.
First, pool size $k$ determines the candidate set passed from Stage~1 to Stage~2;
increasing $k$ raises Stage~2 mAP recall but reduces throughput, so the joint
$(k,\alpha)$ frontier is not separable: optimizing $\alpha$ at a suboptimal $k$
cannot reach the Score achievable by joint search.
Empirically, at $k{=}50$ (QPS\,$\approx$\,295) the SIEVE Score collapses to~0
unless $\alpha$ is simultaneously raised to compensate for reduced Stage-1 recall;
the optimal $k{=}50$, $\alpha{=}0.80$ configuration (Score\,$\approx$\,300)
is reachable only through joint search---independent $\alpha$-optimization at
any fixed $k$ misses it.
Second, engine selection sets the QPS range available to any $\mathit{efSearch}$
sweep, so structural decisions must precede continuous refinement.
These coupled dependencies motivate a phase-partitioned budget that resolves
structural choices before continuous ones.\looseness=-1

\subsection{History Trajectory and Prompt Generation}

\begin{figure*}[t]
  \centering
  \includegraphics[width=0.77\textwidth]{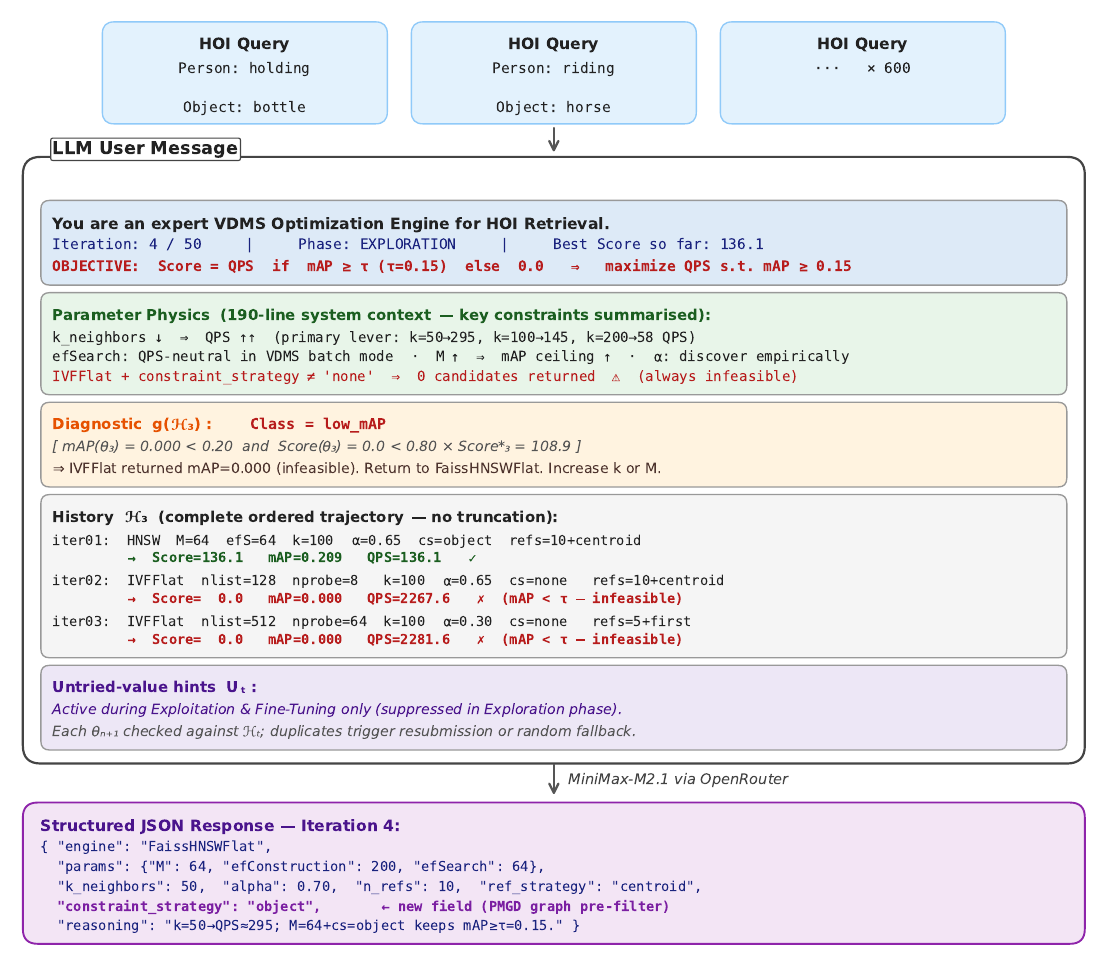}
  \Description{LLM prompt structure: a single user message encodes parameter physics,
    diagnostic class g(H), history, and untried-value hints, eliciting a structured
    JSON configuration response including engine, k-neighbors, alpha, n-refs,
    ref-strategy, constraint-strategy, and reasoning fields.}
  \caption{LLM prompt structure (parameter physics, history $\mathcal{H}_{t-1}$, and phase-conditioned hints).}
  \label{fig:prompts}
\end{figure*}

Let $h_t = \bigl(\theta_t,\,\mathrm{mAP}(\theta_t),\,\mathrm{QPS}(\theta_t)\bigr)$
denote the evaluation record at iteration $t$ and define the
\emph{search history trajectory}
\begin{equation}
  \mathcal{H}_t \;=\; \bigl(h_1,\,h_2,\,\ldots,\,h_t\bigr),
  \quad t \geq 0,\quad \mathcal{H}_0 = \emptyset.
\end{equation}
$\mathcal{H}_t$ is the agent's complete internal state: no surrogate model,
gradient estimate, or kernel is maintained between iterations.
At each iteration the agent receives a single structured prompt built from four
components: (i) fixed system context encoding the VDMS pipeline architecture and
per-engine parameter physics; (ii) the diagnostic class $g(\mathcal{H}_{t-1})$
(Section~\ref{sec:guidance}); (iii) the complete ordered history $\mathcal{H}_{t-1}$
as structured text; and (iv) dynamically computed untried-value hints
(Section~\ref{sec:hints}).
The agent responds with a structured JSON object specifying the next $\theta_t$ and
a \textit{reasoning} field that articulates its rationale (Figure~\ref{fig:prompts}).
The necessity of history conditioning is verified in Section~\ref{sec:ablation}.\looseness=-1

\subsubsection{Diagnostic Guidance Function}
\label{sec:guidance}

Let $\mathrm{Score}^*_{t}$ denote the best Score seen through iteration~$t$.
The diagnostic class
{\footnotesize%
\begin{equation}
  g(\mathcal{H}_{t}) =
  \begin{cases}
    \textit{excellent}   & \text{if } \mathrm{Score}(\theta_{t})
                           \;\geq\; 0.98\;\mathrm{Score}^*_{t},\\[2pt]
    \textit{low\_mAP}    & \text{if } \mathrm{mAP}(\theta_{t}) < 0.20
                           \text{ and } \mathrm{Score}(\theta_{t})
                           < 0.80\;\mathrm{Score}^*_{t},\\[2pt]
    \textit{low\_QPS}    & \text{if } \mathrm{mAP}(\theta_{t}) \geq 0.20
                           \text{ and } \mathrm{QPS}(\theta_{t}) < 100,\\[2pt]
    \textit{below\_best} & \text{if } \mathrm{Score}(\theta_{t})
                           < 0.90\;\mathrm{Score}^*_{t},\\[2pt]
    \textit{moderate}    & \text{otherwise,}
  \end{cases}
  \label{eq:guidance}
\end{equation}%
}
is injected into the prompt as a targeted natural-language directive.
\textit{Excellent} redirects to unexplored structural regions in early phases and
shifts to incremental parameter micro-adjustment in Fine-Tuning;
\textit{low\_mAP} directs the agent to increase $k$;
\textit{low\_QPS} directs it to reduce $k$ (the primary QPS lever) or switch engines;
\textit{below\_best} directs a return toward the best-known configuration with targeted adjustments;
and \textit{moderate} suggests coordinated adjustment of $k$ and $\alpha$.
All threshold values were fixed by empirical calibration on HICO-DET before any optimizer comparison was run.
\subsubsection{Untried-Value Hints and Duplicate Avoidance}
\label{sec:hints}

During EXPLOITATION and FINE-TUNING, the prompt additionally provides
$\mathcal{U}_t$: untried $\alpha$ values and $(n_r, \mathrm{ref\_strategy})$ pairs
under the current best engine, derived from $\mathcal{H}_t$.
These hints prevent the agent from re-proposing already-evaluated
continuous-parameter combinations once the structural choice is settled.
Each proposed $\theta_{t+1}$ is checked against $\mathcal{H}_t$ before evaluation;
if a duplicate is detected, the LLM is re-queried up to four times with the same prompt
(a different response may arise from the model's sampling temperature);
if all four attempts remain duplicates, a uniformly random fallback is applied,
guaranteeing loop progress.
Each LLM call consumes approximately 2{,}000--5{,}000 tokens (growing with history; for larger $N$, top-$k$ history retention is a natural mitigation); the full 50-iteration run costs fewer than 250{,}000 tokens (\textless\$0.20), negligible relative to the 6.5--7.5~h oracle cost.\looseness=-1

\subsection{Phase-Aware Budget Partition}
\label{sec:phases}

The $N$-iteration budget is divided into three phases with boundaries
\begin{align}
  t_{\mathrm{exp}}  &= \operatorname{round}(0.40\,N), \label{eq:texp}\\
  t_{\mathrm{expl}} &= \operatorname{round}(0.75\,N), \label{eq:texpl}
\end{align}
defining \textbf{Exploration} ($1 \leq t \leq t_{\mathrm{exp}}$),
\textbf{Exploitation} ($t_{\mathrm{exp}} < t \leq t_{\mathrm{expl}}$),
and \textbf{Fine-tuning} ($t_{\mathrm{expl}} < t \leq N$) --- for $N{=}50$:
20, 18, and 12 iterations respectively.
The 0.40 boundary allocates sufficient budget to survey all three engine families
before committing to a structural choice; the 0.75 boundary reserves 25\% of the
budget for Fine-Tuning, enough to sweep the 19 discrete $\alpha$ values under
the best-identified engine.
\looseness=-1
The boundaries encode the structural-before-continuous principle: engine and $M$
determine the achievable QPS ceiling and must be resolved before $\alpha$ and $n_r$
are refined---a conditional dependency that Optuna~TPE, which models each parameter
independently via univariate kernel density estimation, cannot exploit.
The benefit of imposing this ordered structure over unphased search is quantified
in the \emph{phase structure} ablation in Section~\ref{sec:ablation}.\looseness=-1

\paragraph{Exploration ($1 \leq t \leq 20$).}
Diversity-focused engine coverage: the \textit{excellent} directive instructs the agent
to sample structurally distinct (engine, $M$, $k$) combinations;
the $\mathcal{U}_t$ hint set is suppressed to prevent premature convergence.
\paragraph{Exploitation ($21 \leq t \leq 38$).}
$\mathcal{U}_t$ is activated; \textit{excellent} shifts to reinforcing the best-engine
region while broadening $(\alpha, n_r)$ coverage on the identified Pareto frontier.
\paragraph{Fine-tuning ($39 \leq t \leq 50$).}
Hints narrow to unexplored reference-strategy variants; proposals are restricted to
coordinate-wise micro-adjustments of the current best
($\mathit{efSearch}$, $n_r$) for latency-targeted gains.

\paragraph{Anti-collapse and stagnation constraints.}
Let $c_t$ count consecutive proposals in which only $\alpha$ changed
relative to the preceding configuration, and let $\iota_t$ count iterations elapsed
without improvement to $\mathrm{Score}^*_t$.
Two hard rules are encoded in the system prompt:
(1) if $c_t \geq 2$, the next proposal must differ structurally---by changing engine,
$M$, $k$, or $n_r$---before any further $\alpha$ refinement;
(2) if $\iota_t \geq 5$, a stagnation escape directive
overrides the current phase instruction and reverts to broad-coverage exploration.
Embedding these rules in the prompt rather than as external rejection filters makes
structural diversity an explicit part of the agent's reasoning: an external
hard-rejection filter would silently discard proposals and retry, consuming LLM
calls without informing the model \emph{why} its proposal was rejected; the
prompt-embedded rule requires the agent to reason about and justify a structurally
distinct alternative.\looseness=-1
Algorithm~\ref{alg:agent} gives the complete procedure.
\begin{algorithm}[t]
\small
\caption{LLM-Guided ANN Optimizer}
\label{alg:agent}
\begin{algorithmic}[1]
\setlength{\itemsep}{-1.5pt}
\Require Configuration space $\Theta$; oracle $\mathcal{O}$; budget $N$
\Ensure Best configuration $\theta^*$ and history $\mathcal{H}_N$
\State $\mathcal{H}_0 \gets \emptyset$;\enspace
       $\mathrm{Score}^* \gets 0$;\enspace
       $\iota \gets 0$
\State $t_{\mathrm{exp}} \gets \operatorname{round}(0.40N)$;\enspace
       $t_{\mathrm{expl}} \gets \operatorname{round}(0.75N)$
       \hfill\Comment{Eqs.~\eqref{eq:texp}--\eqref{eq:texpl}}
\For{$t = 1, 2, \ldots, N$}
  \State $\phi_t \gets \mathrm{Phase}(t,\,t_{\mathrm{exp}},\,t_{\mathrm{expl}})$
         \hfill\Comment{Expl.\ / Exploit.\ / Fine-Tuning}
  \State $g_t \gets g(\mathcal{H}_{t-1})$ \hfill\Comment{Eq.~\eqref{eq:guidance}}
  \State $\mathcal{U}_t \gets \mathrm{UntriedHints}(\mathcal{H}_{t-1},\,\phi_t)$
         \hfill\Comment{$\emptyset$ during EXPLORATION}
  \State $c_t \gets \mathrm{ConsecAlphaOnly}(\mathcal{H}_{t-1})$
         \hfill\Comment{anti-collapse count}
  \For{$\mathit{attempt} = 1,\ldots,4$} \Comment{re-query; temp.\ diversity}
    \State $p_t \gets \mathrm{BuildPrompt}(\phi_t,\,g_t,\,\mathcal{H}_{t-1},\,\mathcal{U}_t,\,c_t,\,\iota)$
    \State $\theta_t \gets \mathrm{LLM}(p_t)$
           \hfill\Comment{structured JSON response}
    \If{$\theta_t \notin \{\theta_s : s < t\}$}
      \State \textbf{break} \hfill\Comment{novel proposal accepted}
    \EndIf
  \EndFor
  \If{$\theta_t \in \{\theta_s : s < t\}$}
         \hfill\Comment{all 4 attempts duplicated}
    \State $\theta_t \gets \mathrm{RandomConfig}(\Theta)$
  \EndIf
  \State $\bigl(\mathrm{mAP}_t,\,\mathrm{QPS}_t\bigr) \gets \mathcal{O}(\theta_t)$
  \State $\mathcal{H}_t \gets \mathcal{H}_{t-1} \cup \{(\theta_t,\,\mathrm{mAP}_t,\,\mathrm{QPS}_t)\}$
  \State $s_t \gets \mathrm{QPS}_t \cdot \mathbf{1}[\mathrm{mAP}_t \geq \tau]$
         \hfill\Comment{SIEVE Score (Eq.~\eqref{eq:score}), $\tau{=}0.15$}
  \If{$s_t > \mathrm{Score}^*$}
    \State $\mathrm{Score}^* \gets s_t$;\enspace
           $\theta^* \gets \theta_t$;\enspace
           $\iota \gets 0$
  \Else
    \State $\iota \gets \iota + 1$
  \EndIf
\EndFor
\State \Return $\theta^*,\;\mathcal{H}_N$
\end{algorithmic}
\end{algorithm}

GP-BO uses \texttt{GPSampler} (Gaussian Process surrogate with Expected Improvement
acquisition~\cite{snoek2012bayesian}) applied to our single-objective Score.
Optuna~TPE fits univariate kernel density estimators over each parameter independently.
Neither baseline encodes the $(k,\alpha)$ coupling or the engine-conditional
structure of $\Theta$; the LLM agent exploits both through its phase-partitioned
prompt and guidance function.
Unlike $\lambda$-Tune~\cite{giannakouris2025lambdatune}, which chains LLM calls to translate a fixed workload description into knob recommendations, our agent maintains a live score trajectory $\mathcal{H}_t$ and adapts its proposals iteration-by-iteration under the SIEVE constraint.\looseness=-1

\subsubsection{Difficulty-Weighted Objective (QDS)}
\label{sec:qds_def}
Because HOI queries vary widely in retrieval difficulty---some verb--object
categories have hundreds of positive images while others have fewer than
five---a uniform mAP treats all queries equally despite this imbalance.
We therefore optionally upweight structurally hard queries via a Query Difficulty Score.
For query $j$: \emph{gallery entropy} $H_j{=}{-}\sum_i p_{ji}\log p_{ji}$
($p_{ji}{=}\mathrm{softmax}(\mathbf{f}_{q_j}^\top\mathbf{f}_{d_i}/\beta)$, $\beta{=}0.01$) measures
the spread of CLIP mass over the gallery;
\emph{centroid proximity} $\mathrm{CPR}_j{=}\mathbf{f}_{q_j}^\top\hat{\boldsymbol{\mu}}_j$
(cosine similarity to the unit-normalised visual centroid $\hat{\boldsymbol{\mu}}_j$)
measures how well the text query aligns with its HOI-class centroid.
We define $\mathrm{QDS}_j{=}\tilde{H}_j{-}\widetilde{\mathrm{CPR}}_j$ (min-max
normalised), weight $w_j{=}1{+}\widetilde{\mathrm{QDS}}_j{\in}[1,2]$, and replace the standard
mAP with $\mathrm{QDS\text{-}mAP}{=}\sum_j w_j\mathrm{AP}_j/\sum_j w_j$ as the
oracle signal.
QDS performance is reported in Table~\ref{tab:optimizer} and Table~\ref{tab:efficiency}.
QDS applies exclusively to HICO-DET: its semantic verb--object category structure
admits per-query difficulty decomposition; GLDv2 (landmark-instance ground truth)
and SIFT1M (exact $k$-NN ground truth) provide no analogous categorical difficulty
signal and retain their standard objectives.
The LLM Agent + QDS best configuration uses FaissIVFFlat ($\mathit{nlist}{=}128$,
$k{=}500$), achieving $1{,}362.0$ QPS on average (3-seed mean).
This throughput is specific to HICO-DET's 47{,}776-image corpus: at this scale
VDMS scans all $\mathit{nlist}$ Voronoi cells during IVFFlat search regardless of
$\mathit{nprobe}$, making IVFFlat equivalent to an exact flat scan over the
quantized index.
With only ${\approx}373$ vectors per cell on average, the total scan is fast enough
to sustain $>1{,}000$ QPS --- a corpus-size effect that disappears at million-vector
scale, where the same VDMS behavior collapses IVFFlat Score to ${\approx}35$
(verified on SIFT1M, which is why IVFFlat is excluded from the SIFT1M search space).
The QDS result therefore characterizes the optimizer's ability to discover
non-obvious high-throughput configurations at 47K scale, not a general IVFFlat
recommendation.\looseness=-1

\section{Experiments}
\label{sec:experiments}

\subsection{Experimental Setup}

\noindent\textbf{Datasets.}
We evaluate on three benchmarks.
\textbf{HICO-DET} (Section~\ref{sec:benchmark}): 47{,}776 images, 600 compositional
text queries, 90{,}641 ground-truth positive pairs --- a stringent testbed combining
fine-grained HOI categories, long-tailed relevance distributions, and a cross-modal
(text-to-image) retrieval setting.
\textbf{GLDv2}~\cite{weyand2020gld}: Google Landmarks Dataset v2, a large-scale
image-to-image retrieval benchmark with 762{,}000 gallery images and 1{,}129 query
images from the retrieval challenge evaluation split, spanning global landmarks, providing
a $16\times$ larger corpus for cross-domain generalization evaluation.
\textbf{SIFT1M}~\cite{jegou2011pq}: a standard ANN benchmark of one million
128-dimensional SIFT feature vectors with 10{,}000 queries and exact nearest-neighbor
ground truth (evaluated at Recall@10), providing a complexity-matched control --- the index space
reduces to a single-stage HNSW problem with no cross-stage coupling --- to test
whether the LLM's advantage is specific to coupled search spaces.
For HICO-DET, image embeddings are 768-dimensional CLIP ViT-L/14 vectors.
For GLDv2, each image is represented by a 768-dimensional CLIP ViT-L/14 embedding
and a 1024-dimensional DINOv2 ViT-L/14-reg4 embedding, both stored in VDMS.
For SIFT1M, raw 128-dimensional SIFT vectors are stored directly in VDMS without
any reranking stage; Score $=$ QPS if Recall@10\,$\geq\tau{=}0.90$, else $0$.

\noindent\textbf{Compared Methods.}
We compare five strategies, each for $N{=}50$ iterations.\looseness=-1

\noindent\textbf{LLM Agent (Ours):} phase-conditioned prompt with full history $\mathcal{H}_{t-1}$ via MiniMax-M2.1/OpenRouter\footnote{Estimated API cost: 50 iterations $\times$ $\approx$5,000 tokens/call $\approx$ 250K tokens total; at MiniMax-M2.1 list pricing this amounts to under \$0.20 per optimizer run.} (seeds: 42, 99, 200). Backbone portability across GPT-4o-mini and Llama-3.3-70B is verified in Section~\ref{sec:ablation}.

\noindent\textbf{GP-BO:} \texttt{optuna.samplers.\allowbreak GPSampler} (Gaussian Process surrogate, Expected Improvement~\cite{snoek2012bayesian}) (seeds: 42, 99, 200).

\noindent\textbf{Optuna~TPE}~\cite{akiba2019optuna}: univariate kernel density estimators (seeds: 42, 99, 200).

\noindent\textbf{Random Search}~\cite{bergstra2012random}: uniform random sampling (seeds: 42, 99, 200).

\noindent\textbf{Grid Search:} lexicographic 1-D sweep (1~run).
As system baselines we include UniIR~\cite{wei2024uniir}, standalone CLIP
ViT-L/14, standalone DINOv2 ViT-L/14-reg4, and CLIP$+$DINOv2 reranking,
all under exact FaissFlat search (Section~\ref{sec:baselines}).\looseness=-1

\noindent\textbf{Metrics.}\label{sec:metrics}
Retrieval quality: mAP, P@10, Recall@10~\cite{manning2008ir}.
Throughput: batch QPS.
Joint optimization objective: Score\,$=$\,QPS if mAP\,$\geq\tau$ else $0$,
$\tau{=}0.15$ (Eq.~\eqref{eq:score});
for SIFT1M, Score\,$=$\,QPS if Recall@10\,$\geq\tau$ else $0$, $\tau{=}0.90$.

\noindent\textbf{Implementation.}
All experiments run on the NCSA Delta GPU cluster (partition \texttt{gpuA40x4}): one
NVIDIA A40 GPU (48 GB VRAM), 16 CPU cores (AMD EPYC 7763), and exclusive node access.
VDMS is deployed as an Apptainer container (v2.2.1); all retrieval and optimization code
is implemented in Python~3.12.
For Optuna, we use the official \texttt{optuna} library~\cite{akiba2019optuna} with
the default TPE sampler.
Each optimization iteration builds a complete ANN index from scratch and issues all
600 queries in a single batch, requiring approximately 8--9 minutes of wall-clock time.
The full 50-iteration budget therefore demands approximately 6.5--7.5 hours of compute
per method, making sample efficiency --- how quickly each method reaches a high-scoring
configuration --- a practically meaningful criterion.
$N{=}50$ covers 0.05\% of $|\Theta|$ (e.g., 63{,}840 HNSW configs on HICO-DET), totals ${\approx}6.5$--$7.5$\,h of wall-clock compute per method, and matches the protocol of related LLM-guided and Bayesian optimization studies~\cite{lao2024gpttuner,yang2024opro}.\looseness=-1

\subsection{System Baselines}
\label{sec:baselines}

\begin{table*}[t]
\centering
\caption{System baselines on HICO-DET (FaissFlat exact search; Score\,$=$\,QPS if mAP\,$\geq\tau$, $\tau{=}0.15$).}
\label{tab:baselines}
\vspace{3pt}
\footnotesize
\setlength{\tabcolsep}{4pt}
\begin{tabular}{p{3.6cm}lccccc}
\toprule
\textbf{Method} & \textbf{$\alpha$} & \textbf{$k$} & \textbf{mAP} & \textbf{P@10} & \textbf{R@10} & \textbf{Score (QPS)} \\
\midrule
\quad CLIP ViT-L/14                          & 0.0 & 500  & 0.20787 & 0.3370 & 0.0967 &  8.21$^\star$ \\
\quad CLIP ViT-L/14                          & 0.0 & 2000 & 0.23978 & 0.3370 & 0.0967 &  1.17 \\
\quad DINOv2 ViT-L/14-reg4                   & 1.0 & 2000 & 0.21655 & 0.3232 & 0.0800 &  1.16 \\
\quad CLIP\,+\,DINOv2 rerank                 & 0.5 & 2000 & \textbf{0.27629} & \textbf{0.3812} & \textbf{0.1138} & 1.15 \\
\midrule
\multicolumn{7}{l}{\footnotesize $^\star$Optimizer reference baseline (SIEVE Score\,=\,8.21, i.e.\ QPS since mAP\,$\geq\tau$); $k{=}500$ trades mAP for higher QPS vs.\ $k{=}2000$.} \\
\bottomrule
\end{tabular}
\end{table*}

\noindent\textbf{Compositional difficulty.}
Even with exhaustive FaissFlat search and single-query DINOv2 reranking ($n_\text{refs}{=}1$),
the best achievable mAP is $0.276$\footnote{With pseudo-relevance feedback ($n_\text{refs}{=}10$),
the FaissFlat oracle reaches mAP $= 0.426$ at $\alpha{=}0.80$, confirming PRF and index
tuning are complementary optimization axes.} --- reflecting fine-grained categories such as
\textit{ride horse}, \textit{sit\_on horse}, and \textit{straddle horse} that share
the same object but differ only in verb.\looseness=-1

\noindent\textbf{UniIR comparison.}
UniIR's fine-tuned CLIP-SF achieves mAP $= 0.095$ in VDMS --- a $2.5\times$ drop
from zero-shot CLIP ($0.240$; Table~\ref{tab:baselines}) --- because multi-task fine-tuning on COCO-style pairs
disrupts HOI semantic alignment for compositional queries.
Under the SIEVE metric, all UniIR variants score $0$ (mAP $< \tau{=}0.15$),
failing to meet the minimum quality threshold; our unoptimized CLIP baseline
already achieves SIEVE Score\,$=$\,8.21 (QPS at $k{=}500$; Table~\ref{tab:baselines}).\looseness=-1

\noindent\textbf{Modality complementarity and throughput.}
DINOv2 alone (mAP\,$=$\,0.217) underperforms CLIP alone (0.240) by 2.3\,pp,
confirming language-aligned embeddings are essential; weighted fusion at $\alpha{=}0.5$
achieves mAP\,$=$\,0.276 ($+$3.65\,pp, $+$15.2\% relative) at only 0.6\% extra latency.
Reducing $k$ from 2000 to 500 raises QPS by $7.0\times$ at 3.2\,pp mAP cost,
yielding a $7.0\times$ Score gain --- the non-linear asymmetry the LLM agent exploits
by conditioning each proposal on the full history of observed outcomes.\looseness=-1

\subsection{End-to-End Optimization Performance}
\label{sec:results_optimizer}

\begin{figure}[t]
  \centering
  \vspace{-10pt}
  \includegraphics[width=\columnwidth]{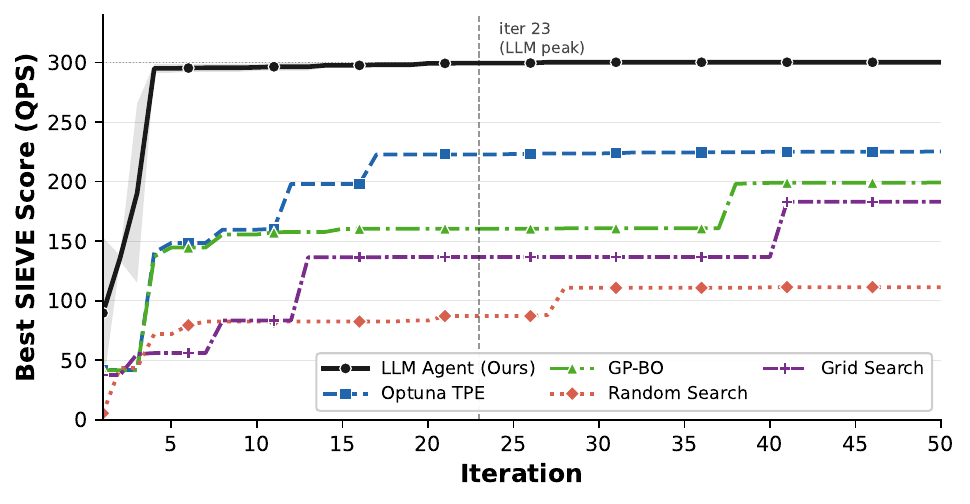}
  \Description{Convergence curves showing best SIEVE Score (QPS if mAP≥τ, τ=0.15) vs.\ iteration
    for five optimizer methods on HICO-DET over 50 iterations. LLM Agent outpaces all baselines
    by iteration 4, reaching a final mean score of 300.3; Optuna~TPE and GP-BO plateau at
    225.2 and 199.3; Grid and Random Search plateau below 184.}
  \vspace{-12pt}
  \caption{SIEVE Score vs.\ iteration on HICO-DET ($\pm 1\sigma$; 3 seeds except Grid).}
  \label{fig:convergence}
\end{figure}

\begin{table*}[t]
\centering
\caption{ANN optimizer comparison on HICO-DET (50~iters, 3-seed means; Grid: 1~run). $^\P$\texttt{GPSampler+EI}~\cite{snoek2012bayesian}; $^\S$difficulty-weighted; $^\dagger$corpus-scale artifact (see Section~\ref{sec:qds_def}); $^{\ddagger}$SIEVE Score\,$=\,0$ (mAP\,$<\tau{=}0.15$); raw QPS\,$=\,19.67$\,/\,$19.46$ respectively.}
\label{tab:optimizer}
\vspace{3pt}
\small
\setlength{\tabcolsep}{6pt}
\begin{tabular}{lcccc}
\toprule
\textbf{Method} & \textbf{Score (QPS)}$\uparrow$ & \textbf{mAP}$\uparrow$ & \textbf{P@10}$\uparrow$ & \textbf{Iters$\to$Best} \\
\midrule
UniIR CLIP-SF Large, image-only~\cite{wei2024uniir} & 0$^{\ddagger}$  & 0.095 & 0.448 & --- \\
UniIR CLIP-SF Large, multi-modal~\cite{wei2024uniir} & 0$^{\ddagger}$  & 0.089 & 0.442 & --- \\
Grid Search                               & 183.0          & 0.178 & 0.508 & 41/50 \\
Random Search~\cite{bergstra2012random}   & 111.4          & 0.179 & 0.455 & 29/50 \\
Optuna~TPE~\cite{akiba2019optuna}         & 225.2          & 0.164 & 0.465 & 41/50 \\
GP-BO$^\P$                                & 199.3          & 0.178 & 0.482 & 39/50 \\
VDTuner~\cite{yang2024vdtuner}            & 223.8          & 0.172 & 0.479 & 24/50 \\
\midrule
\textbf{LLM Agent (Ours)}                 & \textbf{300.3} & 0.161 & 0.438 & 23/50 \\
\textbf{LLM Agent + QDS (Ours)}$^{\S\dagger}$ & \textbf{1362.0}& 0.175 & 0.292 & 39/50 \\
\bottomrule
\end{tabular}
\end{table*}

\begin{table*}[t]
\centering
\caption{System efficiency on HICO-DET (47{,}776 images, 600 queries; best-configuration per representative seed; 3-seed mean Scores in Table~\ref{tab:optimizer}). $^\S$difficulty-weighted; $^\dagger$corpus-scale artifact (see Section~\ref{sec:qds_def}).}
\label{tab:efficiency}
\vspace{3pt}
\small
\setlength{\tabcolsep}{5pt}
\begin{tabular}{lllccccc}
\toprule
\textbf{Configuration} & \textbf{Engine} & \textbf{$k$} & \textbf{Build(s)}$\downarrow$ & \textbf{CLIP(ms)}$\downarrow$ & \textbf{DINO(ms)}$\downarrow$ & \textbf{Lat(ms)}$\downarrow$ & \textbf{QPS}$\uparrow$ \\
\midrule
\multicolumn{8}{l}{\textit{UniIR~\cite{wei2024uniir} ($k{=}10$)}} \\
CLIP-SF Large, image-only         & FaissFlat & 10   & 427.6 & 50.8  & ---   &  50.8 & 19.67 \\
CLIP-SF Large, multi-modal         & FaissFlat & 10   & 427.7 & 51.4  & ---   &  51.4 & 19.46 \\
\midrule
\multicolumn{8}{l}{\textit{Exact-search baselines}} \\
CLIP only ($\alpha{=}0$)          & FaissFlat & 500  & 433.3 & 121.0 & 0.03  & 121.1 & 8.21 \\
CLIP only ($\alpha{=}0$)          & FaissFlat & 2000 & 431.2 & 850.0 & 0.09  & 850.1 & 1.17 \\
DINOv2 only ($\alpha{=}1$)        & FaissFlat & 2000 & 432.9 & 851.6 & 5.18  & 856.8 & 1.16 \\
CLIP$+$DINOv2 ($\alpha{=}0.5$)   & FaissFlat & 2000 & 430.1 & 855.2 & 4.82  & 860.0 & 1.15 \\
\midrule
\multicolumn{8}{l}{\textit{Optimizer best configurations}} \\
Grid Search                  & HNSW $M{=}64$        & 100 &  13.2 &  5.40 & 0.04 &  5.46 &  183.0 \\
Random Search~\cite{bergstra2012random} & HNSW $M{=}8$ & 100 &  11.7 &  6.95 & 0.13 &  7.13 &  140.2 \\
Optuna~TPE~\cite{akiba2019optuna} & HNSW $M{=}16$   &  50 &  13.3 &  3.19 & 0.11 &  3.33 &  300.7 \\
VDTuner~\cite{yang2024vdtuner}   & HNSW $M{=}64$   &  50 &  13.4 &  3.23 & 0.08 &  3.35 &  298.9 \\
\textbf{LLM Agent (Ours)}    & HNSW $M{=}64$        &  50 &  12.6 &  3.21 & 0.08 &  3.33 & \textbf{300.5} \\
\textbf{LLM Agent + QDS (Ours)}$^{\S\dagger}$ & IVFFlat $nl{=}128$ & 500 &  17.2 &  0.35 & 0.34 &  0.74 & \textbf{1357.6} \\
\bottomrule
\end{tabular}
\end{table*}

The LLM agent achieves mean Score $= 300.3$, outperforming every baseline:
Optuna~TPE ($225.2$, $\mathbf{+33.3\%}$), VDTuner~\cite{yang2024vdtuner} ($223.8$, $\mathbf{+34.2\%}$), GP-BO ($199.3$, $\mathbf{+50.7\%}$),
Grid Search ($183.0$, $+64.1\%$), and Random Search ($111.4$).
This corresponds to a $\mathbf{15.3\times}$ throughput gain over UniIR
(QPS\,$=$\,19.67; SIEVE Score\,$=$\,$0$ since mAP\,$<\tau$) and a
$\mathbf{36.6\times}$ gain over the unoptimized CLIP FaissFlat reference
(Score\,$=$\,8.21).
The LLM-optimal configuration --- FaissHNSWFlat, $M{=}64$,
$\mathit{efSearch}{=}64$, $k{=}50$, $\alpha{=}0.80$, $n_r{=}5$,
\textit{centroid} (Table~\ref{tab:efficiency}) --- achieves
QPS\,$=$\,$300.5$ while maintaining mAP\,$=$\,$0.160$ just above the
feasibility threshold $\tau$.
The mAP of $0.160$ falls below the unoptimized CLIP baseline ($0.208$) by design:
SIEVE rewards throughput within the feasible region, so the agent trades excess
mAP margin for higher QPS rather than maximizing mAP.\looseness=-1

\noindent\textbf{GP-BO comparison.}
GP-BO achieves mean Score $= 199.3$, trailing the LLM agent by $50.7\%$,
but with extreme seed-to-seed variance: best-scores span $110$--$299$ across
three seeds versus the LLM agent's $299.9$--$300.5$ ($<0.3\%$ spread).
The cause is structural: GP-BO's smooth Gaussian Process surface cannot model
the discontinuous feasibility boundary at $\tau$ --- when some $k{=}50$ trials
are feasible (Score\,$\approx$\,$300$) and others score $0$ due to an
insufficient $\alpha$, the GP averages these into an expected value of
$\approx\!150$ with high uncertainty, making $k{=}50$ appear less attractive
than the consistently feasible $k{=}100$ region (Score\,$\approx\!145$).
Without natural-language reasoning to diagnose \emph{why} $k{=}50$ fails,
GP-BO cannot locate the joint $(k,\alpha)$ optimum; a constrained-BO variant
would better model the $\tau$ cliff but not the cross-stage $(k,\alpha)$
coupling, which requires joint history-conditioned reasoning.\looseness=-1

\noindent\textbf{Optuna~TPE comparison.}
Optuna~TPE achieves Score $= 225.2$, trailing the LLM agent by
$\mathbf{33.3\%}$ --- the largest consistent gap among adaptive baselines.
The cause is structural: TPE fits \emph{independent} kernel density estimators
over each parameter, so it cannot discount the FaissIVFFlat region without a
joint conditional model over engine type; all three seeds encounter
FaissIVFFlat configurations (mAP\,$=$\,0, Score\,$=$\,0) within the first ten
iterations, biasing the KDE mass toward the infeasible high-throughput region.
The phase-structured budget directly counters this by resolving engine selection
during Exploration (iterations~1--20) before any continuous $(\alpha, n_r)$
refinement --- the same separation whose removal costs $-0.3\%$ in the phase
ablation (Table~\ref{tab:ablation_arch}), confirming it as the mechanism rather
than a coincidence.
As a result, Optuna~TPE's final mean Score ($225.2$) falls below the LLM
agent's iteration-6 Score --- a structural deficit, not a sample-size one.\looseness=-1

\noindent\textbf{Convergence and sample efficiency.}
Figure~\ref{fig:convergence} plots best-Score-so-far mean $\pm 1\sigma$ curves.
The LLM agent reaches Score $= 295.4$ at iteration~6, already exceeding Grid
Search's final mean ($183.0$) and well ahead of Optuna~TPE
(mean\,$=$\,$148.5$) and Random Search ($79.4$) at the same iteration.
A Bartlett variance test ($p{=}1.6{\times}10^{-4}$) confirms the reliability gap:
LLM $\sigma{=}0.38$ versus Optuna~TPE $\sigma{=}65.4$ (bimodal: two seeds at
Score\,${\approx}\,188$, one at ${\approx}\,301$).\looseness=-1

\noindent\textbf{Non-obvious operating point.}
Under the SIEVE objective, the optimal region is $k{=}50$: FaissHNSWFlat
with $k{=}50$ achieves QPS\,$\approx\!300$ but mAP sits at the edge of
feasibility ($\approx\!0.16$, just above $\tau{=}0.15$).
This boundary-adjacent optimum is structurally non-trivial: without the
compensating combination of $\alpha{=}0.80$ and $n_r{=}5$ centroid
references, $k{=}50$ falls below $\tau$ and scores $0$.
The LLM agent locates this joint $(k,\alpha,n_r)$ operating point because
its history-conditioned reasoning explicitly tracks the coupling: after
observing that $k{=}100$ configurations are feasible at moderate $\alpha$,
it probes $k{=}50$ with compensating $\alpha$ --- an inference that Optuna
TPE's independent marginals over $k$ and $\alpha$ cannot make, explaining
why the LLM reaches the $k{=}50$ feasible region on all three seeds
while GP-BO succeeds on only one.
Across all $188$ logged $k{=}50$ configurations, $\alpha{<}0.70$ yields only $46.2\%$ feasibility (mean Score\,$=$\,$129.1$) versus $84.2\%$ at $\alpha{\geq}0.70$ (mean Score\,$=$\,$242.6$) --- a $113$-point gap marginal optimizers cannot reliably navigate.\looseness=-1

\noindent\textbf{Failure modes.}
Grid Search reaches its best Score of 183.0 at iteration~41 --- well below the
LLM optimum (300.3) --- because its one-dimensional lexicographic sweep cannot
jointly probe the $(k{=}50,\,\alpha{=}0.80)$ operating point: sweeping $k$ and
$\alpha$ independently misses the coupled interaction that elevates Score from
183 to 300, even though M\,$=$\,64 is individually explored from iteration~5.\looseness=-1
Random Search's mean Score ($111.4$) reflects frequent infeasible trials
(score\,$=$\,$0$); the feasible configurations it finds tend toward large-$k$
safe regions (Score\,$\approx\!140$) rather than the tight $k{=}50$ optimum
that requires coupled $(k,\alpha)$ inference to locate reliably.\looseness=-1

\subsection{Cross-Domain Generalization: GLDv2 Landmark Retrieval}
\label{sec:gldv2}

We evaluate on GLDv2~\cite{weyand2020gld} ($762{,}000$ gallery images, $1{,}129$ queries) to test cross-domain generalization.
The corpus is $16\times$ larger than HICO-DET, the retrieval task shifts from text-to-image to image-to-image (removing the text-query centroid advantage), and each image is represented by CLIP ViT-L/14 and DINOv2 ViT-L/14-reg4 embeddings in VDMS under the same SIEVE objective (Score $=$ QPS if mAP$\geq\tau{=}0.15$, else $0$).\looseness=-1

\begin{table}[t]
\centering
\caption{Optimizer comparison on GLDv2 (762K gallery, 1{,}129 queries, 50 iterations, 3-seed means; Grid: 1~run).}
\label{tab:gldv2}
\resizebox{\columnwidth}{!}{%
\small
\setlength{\tabcolsep}{4pt}
\begin{tabular}{lccccc}
\toprule
\textbf{Method} & \textbf{Score (QPS)}$\uparrow$ & \textbf{mAP}$\uparrow$ & \textbf{P@10}$\uparrow$ & \textbf{QPS}$\uparrow$ & \textbf{Iters$\to$Best} \\
\midrule
GP-BO~\cite{snoek2012bayesian}  & 56.57  & 0.1892 & 0.2235 & 56.57  & 35/50 \\
Random Search~\cite{bergstra2012random} & 265.47 & 0.1594 & 0.1949 & 265.47 & 30/50 \\
Optuna~TPE~\cite{akiba2019optuna} & 272.35 & 0.1610 & 0.1963 & 272.35 & 37/50 \\
Grid Search                   & 272.98 & 0.1615 & 0.1965 & 272.98 & 6/50  \\
\textbf{VDTuner}~\cite{yang2024vdtuner} & \textbf{273.91} & 0.1593 & 0.1953 & \textbf{273.91} & 43/50 \\
\midrule
LLM Agent (Ours)              & 271.45 & 0.1613 & 0.1961 & 271.45 & 23/50 \\
\bottomrule
\end{tabular}}
\end{table}

\noindent\textbf{Overall performance.}
With 3-seed means, VDTuner leads at $\mathbf{273.91}$ QPS, narrowly ahead of
Grid Search ($272.98$), Optuna~TPE ($272.35$), LLM Agent ($271.45$), and Random Search ($265.47$).
The spread among the top-4 adaptive methods is under $1\%$ --- substantially compressed
relative to HICO-DET ($33.3\%$) --- consistent with the lower coupling on this dataset.
The LLM Agent's fourth-place VDMS result ($271.45$) falls within this $<1\%$ convergence
band and is consistent with 3-seed measurement variance; it is not a structural loss.
GP-BO collapses to a 3-seed mean of $56.6$ QPS, replicating the cliff failure from HICO-DET.
Each iteration completes in $\approx$5\,min ($\approx$4\,h total), slightly faster than HICO-DET.\looseness=-1

\noindent\textbf{GP-BO cliff effect on GLDv2.}
GP-BO achieves a 3-seed mean of $56.6$ QPS (vs.\ $\approx\!273$ for the top methods), the same structural failure as on HICO-DET but more acute.
The SIEVE objective imposes a hard feasibility cliff at $\tau{=}0.15$: configurations with mAP just below this threshold score $0$ regardless of QPS.
GP-BO's smooth Gaussian Process surface cannot model this discontinuity --- it interpolates over the cliff, assigning moderate expected value to boundary-adjacent configurations instead of committing to the high-QPS side.
As a result, GP-BO locked onto $k{=}150$ for $40$ of $50$ iterations (maximizing mAP at the cost of QPS), never probing $k{=}100$ --- a high-scoring setting (mAP$\approx$0.165, QPS$\approx$273) --- because the GP surface predicted it lay in a risky low-mAP zone given nearby $k{=}50$ trials with Score\,$=$\,$0$.\looseness=-1

\noindent\textbf{Sample efficiency.}
Grid Search finds its best Score at iteration~6 (3-seed mean); LLM Agent converges at
iteration~23, VDTuner at iteration~43.
The compressed performance landscape at 762K scale means high-QPS HNSW configurations
are accessible to all methods, and the LLM's history-conditioned reasoning confers no
systematic advantage --- consistent with the complexity-proportional prediction.\looseness=-1

\subsection{Complexity-Matched Validation: SIFT1M}
\label{sec:sift1m}

SIFT1M~\cite{jegou2011pq} ($1$M 128-d SIFT vectors, $10{,}000$ queries) provides a complexity-matched control: the optimizer tunes only $M$, \textit{efConstruction}, \textit{efSearch}, and $k$ --- four uncoupled HNSW parameters with no reranking stage or fusion weight $\alpha$.
We evaluate five methods (LLM Agent, Optuna~TPE, Grid, Random, and GP-BO) under the same 50-iteration budget.\looseness=-1

\begin{table}[t]
\centering
\caption{Optimizer comparison on SIFT1M (1M vectors, 10K queries, 50 iters, 3-seed means; Grid: 1~run; Score\,$=$\,QPS if Recall@10\,$\geq 0.90$).}
\label{tab:sift1m}
\small
\setlength{\tabcolsep}{4pt}
\begin{tabular}{lccc}
\toprule
\textbf{Method} & \textbf{Score (QPS)}$\uparrow$ & \textbf{Recall@10}$\uparrow$ & \textbf{Iters$\to$Best} \\
\midrule
GP-BO~\cite{snoek2012bayesian} & 755.4  & 0.9342 & 18/50 \\
Random Search~\cite{bergstra2012random} & 1143.6 & 0.9268 & 32/50 \\
VDTuner~\cite{yang2024vdtuner} & 1150.9 & 0.9290 & 32/50 \\
Optuna~TPE~\cite{akiba2019optuna} & 1160.7 & 0.9120 & 24/50 \\
Grid Search               & 1174.4 & 0.9146 & 5/50 \\
\midrule
\textbf{LLM Agent (Ours)} & \textbf{1184.5} & 0.9069 & 21/50 \\
\bottomrule
\end{tabular}
\end{table}

\noindent\textbf{Overall performance.}
All methods except GP-BO converge to FaissHNSWFlat with \textit{efSearch}$\in\{24,32\}$
as the optimal operating point. 3-seed mean Scores are
$\mathbf{1184.5}$ (LLM Agent), $1174.4$ (Grid, 1~run), $1160.7$ (Optuna), $1150.9$ (VDTuner), and $1143.6$ (Random) ---
a spread under $3.6\%$.
GP-BO collapses to a 3-seed mean of $755.4$ ($-36.2\%$ vs.\ LLM Agent): its acquisition surface
converges on $k{=}30$ configurations that maximize raw throughput but oscillate around
the $\tau{=}0.90$ recall threshold, with high seed-to-seed variance ($471$--$1174$) and
no recovery across the remaining budget.\looseness=-1

\noindent\textbf{Complexity-proportional advantage.}
The result confirms the paper's central theoretical prediction: the LLM agent's
advantage is complexity-proportional.\looseness=-1
On HICO-DET, the 7-parameter coupled space ($\mathit{efSearch}$--$k$--$\alpha$
interaction across stages) rewards history-conditioned architectural reasoning;
removing the cross-stage coupling, as on SIFT1M, eliminates the advantage because
the residual 4-parameter ANN search space (engine, $M$, $\mathit{efSearch}$, $k$)
is shallow enough for random and grid search to locate the optimum within the same
50-iteration budget.
SIFT1M has no $\alpha$ parameter, so cross-stage coupling is absent by
construction; the remaining $M$--$\mathit{efSearch}$ interaction is monotone
($\rho(M,\text{Score}){=}{+}0.287$; $\rho(\mathit{efSearch},\text{Score}){=}{+}0.243$)
and BO-tractable without joint history-conditioned reasoning.
Notably, sample-efficiency rankings invert: Grid Search reaches its best Score at iteration~5 vs.\ iteration~21 for the LLM agent, confirming that architectural history confers no navigational advantage on the uncoupled 4-parameter space.\looseness=-1

\subsection{Cross-System Generalizability (Milvus)}
\label{sec:milvus}

To verify that results are not VDMS-specific, we replicate all five methods on
Milvus~\cite{wang2021milvus} under identical protocols (50 iterations, 3 seeds,
same dataset splits and $\tau$ thresholds, zero code changes to the optimizer).
VDMS's PMGD backend incurs per-query property-graph overhead that caps HNSW
throughput near $300$--$400$~QPS at 47K-vector scale; Milvus's GPU-accelerated
engine sustains $7{,}000+$~QPS on the same corpus.
Table~\ref{tab:milvus} therefore reports within-system relative rankings ---
the comparison tests optimizer transferability, not cross-backend QPS parity ---
and the LLM agent ranks first on all three datasets across all nine seeds.\looseness=-1

On HICO-DET the margin over VDTuner is $+7.0\%$ (7272 vs.\ 6798).
On GLDv2 the margin widens to $+13.3\%$ (6375 vs.\ 5624): Milvus's ${\approx}20{\times}$
higher absolute QPS at 762K scale reinstates $(k,\alpha)$ coupling absent in the
lower-QPS VDMS parameter landscape.
GP-BO collapses ($-70.5\%$ vs.\ LLM), replicating the cliff failure on VDMS.
On SIFT1M the gap compresses to $+5.8\%$ (41\,847 vs.\ VDTuner's 39\,540),
matching the complexity-proportional scaling seen on VDMS.\looseness=-1

\begingroup
\footnotesize
\setlength{\tabcolsep}{4pt}
\captionof{table}{Cross-system results on Milvus (3-seed means, 50~iters).}
\label{tab:milvus}
\setlength{\topsep}{0pt}\setlength{\partopsep}{0pt}%
\begin{center}
\begin{tabular}{lccc}
\toprule
\textbf{Method} & \textbf{HICO-DET}$\uparrow$ & \textbf{GLDv2}$\uparrow$ & \textbf{SIFT1M}$\uparrow$ \\
\midrule
Random Search~\cite{bergstra2012random} & 4504          & 2632          & 31{,}134 \\
Optuna~TPE~\cite{akiba2019optuna}       & 5368          & 4792          & 35{,}330 \\
GP-BO~\cite{snoek2012bayesian}          & 5449          & 1882          & 35{,}051 \\
VDTuner~\cite{yang2024vdtuner}          & 6798          & 5624          & 39{,}540 \\
\midrule
\textbf{LLM Agent (Ours)} & \textbf{7272} & \textbf{6375} & \textbf{41{,}847} \\
\bottomrule
\end{tabular}
\end{center}
\endgroup

\vspace{4pt}
\subsection{Ablation Study}
\vspace{-1pt}
\label{sec:ablation}
\noindent\textbf{Parameter ablations ($\alpha$, $k$, $n_r$, engine).}
$\alpha{=}0.5$ raises mAP from $0.240$ to $0.276$ (\textbf{$+$3.65\,pp}) at $0.6\%$ extra latency; $\alpha{\in}(0,1)$ strictly dominates both pure encoders.
Increasing $k$ $500{\to}2000$ improves mAP but collapses Score $8.21{\to}1.17$ (\textbf{$7{\times}$ throughput penalty}); the agent discovers the non-obvious optimum \textbf{$k{=}50$, $\alpha{=}0.80$, $n_r{=}5$} centroid (Score\,$\approx\!300$, mAP\,$=$\,$0.160$, just above $\tau$).
The \textit{centroid} reference strategy (minimizes $\varepsilon_j$, Eq.~\eqref{eq:imagequery}) outperforms \textit{first} and \textit{diverse}; the agent selects \textbf{HNSW $M{=}64$} on all seeds while Optuna spreads over $M{\in}\{16,32,48\}$ (mean $225.2$), confirming the value of joint engine-type conditioning.\looseness=-1

\noindent\textbf{History conditioning and phase structure.}

\begin{figure}[!h]
  \centering
  \vspace{-12pt}
  \includegraphics[width=\columnwidth]{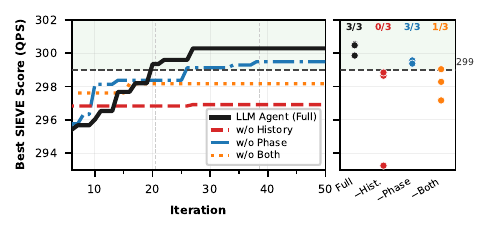}
  \Description{Ablation convergence curves. Without history conditioning the agent
    stagnates completely after iteration 5, gaining only 0.1 Score points over the
    next 45 iterations and reaching Score>=299 in 0 of 3 seeds. The full agent
    improves from 296.0 at iteration 10 to 300.3 at iteration 50, reaching 299+
    on all 3 seeds. Without phase structure the agent still learns (+1.3 gain)
    and reaches 299+ on 3/3 seeds but converges later.}
  \vspace{-14pt}
  \caption{Ablation: Best-Score-so-Far (BSF) convergence (left) and per-seed final Score at $N{=}50$ (right).}
  \label{fig:ablation_convergence}
\end{figure}

Figure~\ref{fig:ablation_convergence} and Table~\ref{tab:ablation_arch} isolate each architectural component across 3 seeds, revealing that the
primary effect of history conditioning is \emph{not} a marginal Score improvement but a binary reliability switch:
removing it causes complete stagnation after iteration~5 ($+0.1$ pts gain over 45 remaining iterations)
and \textbf{0/3 seeds} reach Score\,$\geq$\,299, versus \textbf{3/3} for the full agent ($\sigma$: $3.2\to0.4$).
Without the $(\theta,\mathrm{Score})$ trajectory the agent cannot commit to the $(k{=}50,\,\alpha{=}0.80)$
joint operating point, mirroring the GP-BO cliff failure of Section~\ref{sec:results_optimizer}.\looseness=-1
\emph{\textbf{Phase structure}} has a secondary effect: removing it reduces the mean to $299.5$ ($-0.3\%$) but \textbf{3/3 seeds}
still cross the threshold, confirming that phase ordering accelerates convergence rather than enabling it (and upper-bounding OPRO~\cite{yang2024opro}, which retains history but lacks both).
The non-monotonicity (w/o both: \textbf{$298.2$}, $1/3{\geq}299$ $>$ w/o history alone: \textbf{$296.9$}, $0/3{\geq}299$)
arises because history without phase boundaries anchors the agent to its plateau, while removing both
restores structural exploration that occasionally reaches Score\,$\geq$\,299.\looseness=-1

\begingroup
\small
\setlength{\tabcolsep}{4pt}
\captionof{table}{Mechanistic ablations on HICO-DET (3-seed means, 50~iters).}
\label{tab:ablation_arch}
\vspace{-2pt}
\resizebox{\columnwidth}{!}{%
\begin{tabular}{lcccccc}
\toprule
\textbf{Method} & \textbf{Score@10} & \textbf{Score@50}$\uparrow$ & \textbf{Gain} & \textbf{$\sigma$} & \textbf{Seeds\,$\geq$\,299} & \textbf{$\Delta$@50} \\
\midrule
LLM Agent (Full)               & 296.0 & \textbf{300.3} & $+4.3$ & 0.4 & \textbf{3/3} & --- \\
\quad w/o History conditioning  & 296.8 & 296.9          & $+0.1$ & 3.2 & 0/3          & $-1.1\%$ \\
\quad w/o Phase structure       & 298.1 & 299.5          & $+1.3$ & 0.1 & 3/3          & $-0.3\%$ \\
\quad w/o Both                  & 297.6 & 298.2          & $+0.5$ & 0.9 & 1/3          & $-0.7\%$ \\
\bottomrule
\end{tabular}}
\endgroup
\vspace{4pt}

\noindent\textbf{Backbone portability.}
A cross-backbone check confirms that the phase-aware protocol is model-agnostic:
GPT-4o-mini (seed~42, Score\,$=$\,\textbf{297.44}) and Llama-3.3-70B (Score\,$=$\,\textbf{294.20}) fall within
\textbf{0.95\%} and \textbf{2.0\%} of the MiniMax-M2.1 mean ($300.3$), respectively.\looseness=-1

\noindent\textbf{Phase boundary sensitivity.}
To assess sensitivity to the phase-boundary thresholds $(t_{\mathrm{exp}},t_{\mathrm{expl}})$, we ran two additional configurations at seed~42: equal-thirds ($0.33/0.66$, Score\,$=$\,299.73) and late-exploitation ($0.50/0.80$, Score\,$=$\,302.12), versus the default ($0.40/0.75$, Score\,$=$\,300.54).
All three fall within $\pm$0.53\% of the default, confirming that the optimizer is insensitive to moderate boundary shifts and that the phase-structure benefit reported above holds across schedule variants.\looseness=-1

\section{Conclusion and Outlook}
\label{sec:conclusion}

Two-stage ANN retrieval pipelines impose conditional dependencies ---
engine-type gating, cross-stage search-depth--fusion-weight coupling, and a
discontinuous feasibility cliff --- that parameter-independent HPO methods
model incorrectly by treating jointly coupled parameters as independent
marginals.
We introduced SIEVE, a quality-gated throughput objective, and a phase-aware
LLM agent (Exploration $\to$ Exploitation $\to$ Fine-tuning) that navigates
coupling by conditioning each proposal on the full accumulated optimization history.\looseness=-1

On HICO-DET the agent outperforms the best non-LLM baseline by $+33.3\%$ and achieves 3/3-seed
reproducibility (vs.\ 1/3 for Optuna~TPE and GP-BO); on GLDv2 and SIFT1M performance converges (adaptive methods except GP-BO within $1\%$ and $3.6\%$ respectively),
confirming that \emph{LLM optimization advantage scales with configuration-space coupling complexity}.
Future work includes scaling to billion-vector corpora and extending the
phase-aware framework to multi-objective settings where latency, throughput,
and quality targets are jointly negotiated.\looseness=-1

\vspace{-1pt}
\begin{acks}
This work used NCSA Delta GPU at NCSA through allocation CIS240646 from the Advanced Cyberinfrastructure
Coordination Ecosystem: Services \& Support (ACCESS)
program~\cite{boerner2023access}, which is supported by National Science Foundation grants \#2138259, \#2138286, \#2138307, \#2137603,
and \#2138296.
\end{acks}
\bibliographystyle{abbrv}
\bibliography{sample}

\end{document}